\documentclass[11pt,a4paper]{article}
\usepackage[table]{xcolor}
\usepackage{booktabs} % 专业表格线
\usepackage{array}    % 列格式调整
\usepackage{graphicx} % 表格缩放
\usepackage{caption}  % 表格标题
\usepackage{siunitx}
\usepackage{times,latexsym}
\usepackage{url}
\usepackage[T1]{fontenc}
\usepackage{multirow}
\usepackage{makecell}  % 如果使用makecell方案
\usepackage{tabularx} 
\usepackage{float}
\usepackage{amsmath}  % Required for aligned/gathered environments

\usepackage{colortbl}

\definecolor{lightRed}{RGB}{255, 110, 110}    % 浅红色
\definecolor{lightBlue}{RGB}{110, 110,255}   % 浅蓝色
\definecolor{pureWhite}{RGB}{255, 255, 255}   % 纯白色

\usepackage[table]{xcolor}

\usepackage{pgf} % for calculations

\usepackage{tacl2021v1}
\usepackage{xspace,mfirstuc,tabulary}

\newif\iftaclinstructions
\taclinstructionsfalse % AUTHORS: do NOT set this to true
\iftaclinstructions
\renewcommand{\confidential}{}

\newcommand{\instr}
\fi

\iftaclpubformat % this "if" is set by the choice of options
\newcommand{\taclpaper}{final version\xspace}

\else
\newcommand{\taclpaper}{submission\xspace}

\fi

\title{Revealing the Numeracy Gap: An Empirical Investigation of Text Embedding Models}

\author{
  Ningyuan Deng, Hanyu Duan, Yixuan Tang, Yi Yang \\
  The Hong Kong University of Science and Technology \\
  \texttt{ningyuandeng@ust.hk, hduanac@connect.ust.hk} \\
  \texttt{ytangch@connect.ust.hk, imyiyang@ust.hk} \\
}

\date{}

\begin{document}
\maketitle
\begin{abstract}

Text embedding models are widely used in natural language processing applications. 
However, their capability is often benchmarked on tasks that do not require understanding nuanced numerical information in text. 
As a result, it remains unclear whether current embedding models can precisely encode numerical content, such as numbers, into embeddings. 
This question is critical because embedding models are increasingly applied in domains where numbers matter, such as finance and healthcare. 
For example, ``Company X's market share grew by 2\%'' should be interpreted very differently from ``Company X's market share grew by 20\%,'' even though both indicate growth in market share. 
This study aims to examine whether text embedding models can capture such nuances. 
Using synthetic data in a financial context, we evaluate 13 widely used text embedding models and find that they generally struggle to capture numerical details accurately. 
Our further analyses provide deeper insights into embedding numeracy, informing future research to strengthen embedding model-based NLP systems with improved capacity for handling numerical content.

\end{abstract}

\iftaclpubformat
\section{Courtesy warning: Common violations of \taclpaper rules that have
resulted in papers being returned to authors for corrections}

Avoid publication delays by avoiding these.
\begin{enumerate}
\item Violation: incorrect parentheses for in-text citations.  See \S
\ref{sec:in-text-cite} and Table \ref{tab:cite-commands}.
\item Violation: URLs that, when clicked, yield an error such as a 404 or go
to the wrong page.
  \begin{itemize}
     \item Advice: best scholarly practice for referencing URLS would be to also
     include the date last accessed.
  \end{itemize}
\item Violation: non-fulfillment of promise from submission to provide access
instructions (such as a URL) for code or data.
\item Violation: References incorrectly formatted (see \S\ref{sec:references}).
Specifically:
\begin{enumerate}
  \item Violation: initials instead of full first/given names in references.
  \item Violation: missing periods after middle initials.
  \item Violation: incorrect capitalization.  For example, change ``lstm'' to
  LSTM and ``glove'' to GloVe.
  \begin{itemize}
    \item Advice: if using BibTex, apply curly braces within the title field to
    preserve intended capitalization.
  \end{itemize}
  \item Violation: using ``et al.'' in a reference instead of listing all
  authors of a work.
  \begin{itemize}
    \item Advice: List all authors and check accents on author names even when
    dozens of authors are involved.
  \end{itemize}
  \item Violation: not giving a complete arXiv citation number.
  \begin{itemize}
     \item Advice: best scholarly practice would be to give not only the full
     arXiv number, but also the version number, even if one is citing version 1.
  \end{itemize}
  \item Violation: not citing an existing peer-reviewed version in addition to
  or instead of a preprints
    \begin{itemize}
     \item Advice: When preparing the camera-ready, perform an additional check
     of preprints cited to see whether a peer-reviewed version has appeared
     in the meantime.
  \end{itemize}
  \item Violation: book titles do not have the first initial of all main words
  capitalized.
  \item Violation: In a title, not capitalizing the first letter of the first word
  after a colon or similar punctuation mark.
\end{enumerate}
\item Violation: figures or tables appear on the first page.
  \item Violation: the author block under the title is not
    formatted according to Appendix~\ref{sec:authorformatting}.
\end{enumerate}
\else
% Submission-specific rules
% \section{q}
% 1.现在200的dataset，做成700test有用嘛
% 2. 现在只用了一个domain，扩展到其他domain.
\section{Introduction}

%Text embeddings transform words and sentences into dense vectors that machines can understand and compare. These representations have become essential infrastructure for modern NLP systems; they power semantic search, enable retrieval-augmented generation, and support countless downstream applications. Recent advances, particularly those leveraging large language models like E5-Mistral-7B \citep{Wang2023ImprovingTE}, have achieved impressive performance on standard benchmarks including MTEB \citep{muennighoff-etal-2023-mteb} and BEIR \citep{Thakur2021BEIRAH}. Despite these successes, a fundamental question remains about the embedding models' capacity to handle numerical relationships. While embeddings excel at capturing semantic similarities, their ability to preserve and compare numerical magnitudes remains inadequately understood. This is particularly critical in domains like finance, medicine, and scientific applications, where numerical precision matters. Thus, understanding whether and how well embeddings encode numerical relationships is crucial for these applications, yet this aspect remains largely unexplored in current literature.

Text embedding models transform words and sentences into dense vectors that machines can understand and compare~\citep{zhang2025role}.
These vectors have become essential infrastructure for modern NLP systems; they power semantic search \citep{muennighoff2022sgpt}, enable retrieval-augmented generation (RAG) \citep{fan2024survey}, and support countless downstream applications. 
Recent advances, particularly those leveraging large language models like E5-Mistral-7B \citep{Wang2023ImprovingTE}, have achieved impressive performance on standard benchmarks including MTEB \citep{muennighoff-etal-2023-mteb} and BEIR \citep{Thakur2021BEIRAH}. 
Despite these successes, current benchmarks often overlook evaluating embedding models' capability to handle nuanced numerical content in text. 
Yet, there is a clear need to assess how well embedding models capture numerical details, especially in areas like finance, healthcare, and scientific research, where numerical precision matters. 

For example, consider clinical triage notes that record a patient's blood pressure. Sentences such as ``blood pressure is 120/80 mmHg'' and ``blood pressure is 180/110 mmHg'' are nearly identical in wording, yet they indicate very different levels of risk. When these notes are transformed into embeddings to support clinical automation, the numeric values must be faithfully preserved so that the system can distinguish normal readings from dangerous ones. Without this capability, embeddings may not be able to support, or could even mislead, decision-making in essential clinical tasks, such as alerting clinicians when a patient's condition exceeds safe thresholds.

This work aims to fill this evaluation gap. To support the evaluation, we introduce $\mathtt{EmbedNum}$-$\mathtt{1K}$, a financial domain-specific dataset designed to test whether numerical information in text is sufficiently preserved in embeddings. 
The dataset contains 1,000 synthetic samples, each comprising a question ($\mathbf{Q}$) and two candidate answers ($\mathbf{A^+}$ and $\mathbf{A^-}$). For example, a sample may look like:

\begin{itemize}
    \item \textbf{$\mathbf{Q}$:} ``Who owns over 15\% of the company?''
    \item \textbf{$\mathbf{A^+}$:} ``Investor Alice owns a 20\% stake.''
    \item \textbf{$\mathbf{A^-}$:} ``Investor Alice owns a 5\% stake.''
\end{itemize}
Given that each pair of candidate answers differs only in their numeric values, we would expect that an ideal embedding model should encode the question ($\mathbf{Q}$) closer to the correct answer ($\mathbf{A^+}$) than to the incorrect one ($\mathbf{A^-}$) in the embedding space. 
This reflects the fact that 20\% satisfies the condition of ``over 15\%,'' whereas 5\% does not. 
This task resembles a classic retrieval problem in NLP, but with a critical uniqueness: correctness is determined by the numerical details rather than by overall semantic meaning.
Thus, achieving higher accuracy\footnote{Accuracy on this task is calculated as the proportion of samples where the model correctly ranks $\mathbf{A^+}$ closer to $\mathbf{Q}$ than $\mathbf{A^-}$.} on this task indicates that the embedding model more reliably preserves numerical details together with their surrounding context.

Using $\mathtt{EmbedNum}$-$\mathtt{1K}$, we experiment with 13 widely used text embedding models, including BERT-like ones, LLM-based ones, and general-purpose and financial-domain specific ones, either open-source or via commercial API calls.
To enable fine-grained evaluation, we vary the numeric format of numbers.
For example, numbers can appear in different formats, such as integers (e.g., 6), decimals (e.g., 0.6), percentages (e.g., 6\%), and written numbers (e.g., ``six'').

We find that embedding models generally struggle to capture numerical details in text, with accuracy slightly better than chance on average. Notably, LLM-based embedding models show a clear advantage over encoder-based ones. In addition, models interpret numbers differently depending on their numeric format; for example, values such as 8\% and 0.08 are treated distinctly. Furthermore, our closer examination reveals that improved textual literacy in models does not necessarily translate into better numeracy, and out-of-vocabulary (OOV) numbers pose a particular challenge. Finally, and interestingly, we show that just as humans face higher cognitive load and are more prone to errors when processing numbers with many digits, embedding models also struggle with long, high-precision numbers. 

Later in this paper, we conduct additional analyses to gain deeper insights into embedding numeracy. 
We examine how embedding models handle different types of numeric reasoning, compare numeric values in digit form (e.g., ``24'') versus written form (e.g., ``twenty-four''), and assess the impact of surrounding linguistic context, to see whether it helps or weakens the model's ability to capture precise numerical information. 
We also investigate whether frequent exposure to numbers during training helps gain numeracy and evaluate whether probing test results reliably predict numeracy in downstream tasks. 
Together, these analyses provide key insights for improving embedding models' numerical capabilities.

This work makes three main contributions.

First, we introduce $\mathtt{EmbedNum}$-$\mathtt{1K}$, a dataset that fills an important evaluation gap by testing how well current embedding models preserve subtle numerical differences in embeddings, which are often overlooked in existing benchmarks. For example, current benchmarks may treat the embeddings of ``Investor Alice owns a 20\% stake.'' and ``Investor Alice owns a 5\% stake.'' as equally good matches to the embedding of ``Who owns over 15\% of the company?''. In contrast, our evaluation explicitly differentiates such cases, testing whether embeddings capture the precise numerical relationships implied in the text.

Second, we reveal a key limitation of current embedding models: they often fail to accurately preserve subtle numerical information in their embeddings. This suggests that simply scaling up model size or training data, as is common in current embedding model training practices, is not sufficient. Targeted designs that explicitly account for fine-grained information, such as the numerical details studied here, are also necessary.

Third, our experiments generate key insights that may help to develop numeracy-aware text embedding models in the future. We believe this has important implications for advancing number-intensive NLP applications, such as RAG-based systems in finance and healthcare \citep{wong2025position, yepes2024financial}, where accurate numerical understanding is critical. We will make the datasets and experimental code publicly available to ensure reproducibility.

\section{Related Work}

\subsection{Benchmarking Embedding Models and the Evaluation Tasks} 
Numerous benchmarks have been introduced to evaluate embedding models. A well-known example is the Massive Text Embedding Benchmark (MTEB) \citep{muennighoff-etal-2023-mteb}. 
Others include, but are not limited to, MMTEB \citep{Enevoldsen2025MMTEBMM}, which expands the original MTEB to cover more languages, and FinMTEB \citep{Tang2025FinMTEBFM}, which adapts MTEB for domain-specific settings such as finance. 
These benchmarks typically include a broad set of evaluation tasks for embedding models to perform. 
Generally, models more capable of encoding similar texts into similar embeddings tend to achieve higher scores on these tasks.

Although current benchmarks have achieved great diversity in evaluation tasks, it remains unclear how well embedding models handle cases where numerical understanding matters.
Case in point, consider clustering research papers on arXiv\footnote{\url{https://arxiv.org/}} into categories such as Physics and Computer Science (based on their abstracts), a task included in MTEB \citep{muennighoff-etal-2023-mteb}. In this case, the clustering results are largely driven by the overall topic and keyword patterns in the text, while numerical details, such as years, equation numbers, or statistics mentioned in the abstract, play little role in determining the results.

Even in FinMTEB \citep{Tang2025FinMTEBFM}, a recently developed benchmark tailored to the financial domain where numbers clearly play a critical role, the evaluation of embedding models' ability to deal with nuanced numerical content is not explicitly considered. 
Our work contributes to this line of research by highlighting this important evaluation dimension and curating a dataset to assess how well embedding models capture, preserve, and interpret numerical details in text.

\subsection{Numeracy in Embeddings}
Numeracy refers to the ability to understand and work with numbers. 
In NLP, interest in whether embeddings capture numeracy can be traced back as early as 2019, with pioneering works by \citet{naik2019exploring} and \citet{wallace2019nlp}. 
At that time, the focus was primarily on word embeddings such as GloVe \citep{pennington2014glove} and word2vec \citep{mikolov2013efficient}. 
Later, several follow-up works developed approaches that produce number embeddings with improved numeracy preservation \citep{sundararaman2020methods, duan2021learning, jiang2020learning, sivakumar2025leverage}.

Our work falls within this line of research examining numeracy in embeddings and extends it by focusing on large-scale modern text embedding models, such as the Qwen and Mistral embedding series \citep{qwen3embedding, li2023towards, LinqAIResearch2024, SFRAIResearch2024}, going beyond traditional word embeddings. This extension is meaningful in the following two key aspects. 

First, the interpretation of a number in text can vary greatly depending on its surrounding linguistic context. 
For instance, investors may respond very differently to ``Stock A \textit{rose} by 2\%'' versus ``Stock A \textit{fell} by 2\%,'' even though the magnitude of the change is the same.
Additionally, a model's ability to understand numbers can, in turn, enhance its interpretation of the surrounding context \citep{thawani2021numeracy}. 
Unlike word embedding models, which typically encode numbers in a context-free manner, text embedding models enable the examination of numeracy in context (i.e., context-dependent) by embedding entire sentences or paragraphs that contain numbers. 
Given that real-world NLP applications rarely handle numbers in isolation, our examination of numeracy in context enables a more realistic setting.

Second, compared to traditional word embedding models, text embedding models (often powered by LLMs) serve more prominently as core components in modern NLP systems, especially in RAG settings \citep{fan2024survey}.
Therefore, we argue that investigating numeracy in text embedding models is of greater practical relevance.

\section{Dataset Construction}
We describe the curation of $\mathtt{EmbedNum}$-$\mathtt{1K}$ in the following four steps; the overall workflow is illustrated in Figure~\ref{fig: data}.

\begin{figure*}[htbp] 
\centering 
\includegraphics[width=1\linewidth]{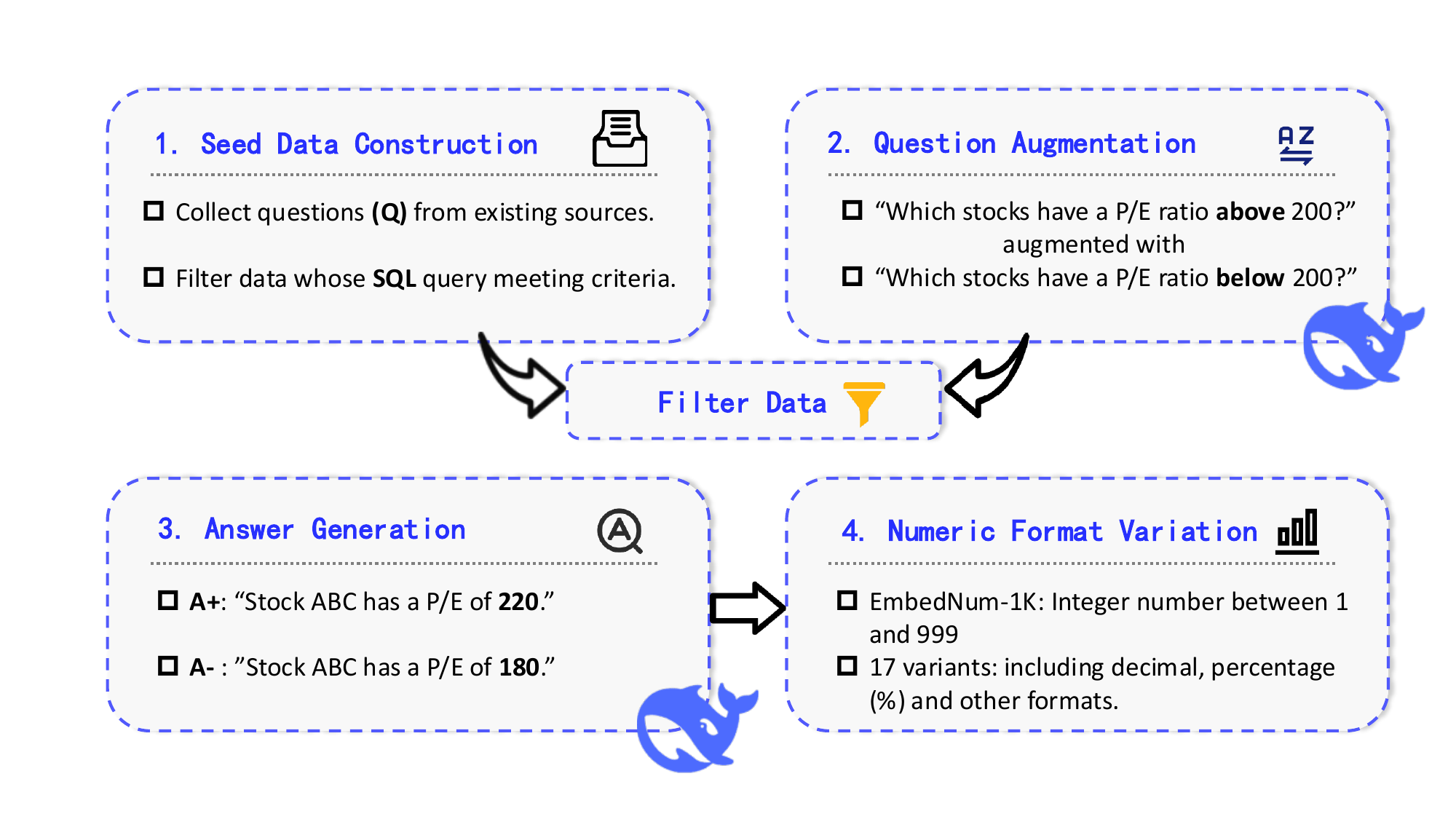} 
\caption{Workflow of the $\mathtt{EmbedNum}$-$\mathtt{1K}$ dataset construction process.} 
\label{fig: data}
\end{figure*}

\noindent \textbf{Step 1: Seed Data Construction.} 
We construct our seed data by filtering the English portion of the BULL dataset.\footnote{\url{https://bull-text-to-sql-benchmark.github.io}} 
This dataset consists of 4,966 natural language question-SQL pairs collected from real-world financial industry settings and was originally curated for financial Text-to-SQL benchmarking \citep{Zhang2024FinSQLML}. 
A question-SQL pair may look like: \textbf{Question:} ``Which stocks have a P/E ratio above 200?'' \textbf{SQL:} ``select chinameabbr from qt\_monthdata where pettm > 200;''.

Here, we are interested only in the \textbf{Question} entries, specifically those whose \textbf{SQL} query counterparts meet the following three criteria: 
(1) the query includes either the ``>'' or ``<'' operator, 
(2) it does not reference any time-related fields,\footnote{Time (date) comparisons are excluded due to their higher complexity.} and 
(3) it contains exactly one numeric value. This yields a total of 545 questions.

\noindent \textbf{Step 2: Question Augmentation.} 
To expand the seed set, for each question obtained from Step 1, we create a paired variant. 
For example, the question ``Which stocks have a P/E ratio \textit{above} 200?'' could be augmented with ``Which stocks have a P/E ratio \textit{below} 200?''. 
This flips the numerical comparison in the original question while keeping the context unchanged. 
We approach this by prompting the DeepSeek-V3.1 model \citep{deepseekai2024deepseekv3technicalreport}\footnote{\url{https://api-docs.deepseek.com}} in non-thinking mode. 
%Then, from the augmented data pool, we randomly sample 500 sentences to get 1,000 questions, while ensuring a balanced distribution between the ``above'' and ``below'' cases.
Then, from the augmented data pool, we randomly select 500 sentences to create 1,000 questions,  while ensuring a balanced distribution between the ``above'' and ``below'' cases.

\noindent \textbf{Step 3: Answer Generation.} 
For each of the 1K questions ($\mathbf{Q}$), we create two candidate answers (by again calling DeepSeek-V3.1 \citep{deepseekai2024deepseekv3technicalreport}): $\mathbf{A^+}$, which satisfies the condition stated in the question $\mathbf{Q}$, and $\mathbf{A^-}$, which differs from $\mathbf{A^+}$ only in its numeric value and does not satisfy the condition. 
For example, take \textbf{$\mathbf{Q}$}: ``Which stocks have a P/E ratio above 200?''; the candidate answers could be: 
\begin{itemize}
    \item \textbf{$\mathbf{A^+}$:} ``Stock ABC has a P/E of \textit{220}.''
    \item \textbf{$\mathbf{A^-}$:} ``Stock ABC has a P/E of \textit{180}.''
\end{itemize}

Up to this point, we have created the original version of $\mathtt{EmbedNum}$-$\mathtt{1K}$, comprising 1K such <$\mathbf{Q}$, $\mathbf{A^+}$, $\mathbf{A^-}$> triples, where all numeric values involved are integers ranging from 1 to 999. 

\noindent \textbf{Step 4: Numeric Format Variation.} 
Following the approach described in \citep{Sivakumar2023FERMATAA}, we construct 17 additional variants of the original $\mathtt{EmbedNum}$-$\mathtt{1K}$ (numbers as integers) by changing the format in which numbers are presented.
These variants include numbers expressed as decimals (e.g., 0.5), percentages (e.g., 5\%), and other formats. 
A detailed description of all variants is provided in Table \ref{tab:data_types}. 

\begin{table}[h]
\centering
\footnotesize
\resizebox{0.475\textwidth}{!}{%
\begin{tabular}{|>{\centering\arraybackslash}m{2.4cm}|>{\raggedright\arraybackslash}m{4.5cm}|>{\raggedright\arraybackslash}m{1.7cm}|}
\hline
\textbf{Format Category} & \multicolumn{1}{c|}{\textbf{Description}} & \multicolumn{1}{c|}{\textbf{Example}} 

\\
\hline
Orig & Integer ranging from 1 to 999 & 235 \\
\hline
\multirow{4}{*}{\makecell[c]{(a) Integers}}
    & (a1) Single-digit integer & 8 \\
    \cline{2-3}
    & (a2) Two-digit integer & 77 \\ 
    \cline{2-3}
    & (a3) Three-digit integer & 719 \\
    \cline{2-3}
    & (a4) Four-digit integer & 3035 \\
\hline
\multirow{4}{*}{\makecell[c]{(b) Decimals}}
    & (b1) One decimal place & 8.3 \\
    \cline{2-3}
    & (b2) Two decimal places & 8.67 \\ 
    \cline{2-3}
    & (b3) Three decimal places & 8.868 \\
    \cline{2-3}
    & (b4) Four decimal places & 8.7713 \\
\hline
\multirow{3}{*}{\makecell[c]{(c) Scaled Down}}
    & (c1) 1-digit scaled down by 10 & 0.8 \\
    \cline{2-3}
    & (c2) 1-digit scaled down by 100 & 0.08 \\
    \cline{2-3}
    & (c3) 1-digit scaled down by 1000 & 0.008 \\
\hline
\multirow{3}{*}{\makecell[c]{(d) Scaled Up}}
    & (d1) 1-digit scaled up by 10 & 80 \\
    \cline{2-3}
    & (d2) 1-digit scaled up by 100 & 800 \\
    \cline{2-3}
    & (d3) 1-digit scaled up by 1000 & 8000 \\
\hline
\multirow{3}{*}{\makecell[c]{(e) Others}}
    & (e1) Written with ``percentage'' & 8 percentage \\
    \cline{2-3}
    & (e2) Written with \% symbol & 8\% \\ 
    \cline{2-3}
    & (e3) Comma-separated integer & 3,035 \\ 
\hline
\end{tabular}
}
\caption{Descriptions of numeric formats used in the experiments with examples.}
\label{tab:data_types}
\end{table}

Altogether, we obtain 18 datasets, including the original $\mathtt{EmbedNum}$-$\mathtt{1K}$ and its 17 variants, which represent numbers in different formats.
The prompts used throughout the data curation process are provided in Appendix~\ref{appx:dataconstruction}. 

\section{Experiments}
\subsection{Experimental Setup}
\noindent \textbf{Task Description.} 
Based on the $\mathtt{EmbedNum}$-$\mathtt{1K}$ dataset, we formulate a retrieval task. Specifically, given an instance <$\mathbf{Q}$, $\mathbf{A^+}$, $\mathbf{A^-}$> from the dataset and a chosen embedding model, the aim of the task is to retrieve the correct answer $\mathbf{A^+}$ from the two candidate answers based on the question $\mathbf{Q}$. To do this, we first encode the question and both candidate answers into dense embedding vectors using the model under evaluation. We then compute the cosine similarity between the question embedding and each of the two candidate answer embeddings, yielding one similarity score per answer. The retrieved answer is then determined as the one with the higher similarity score.

\noindent \textbf{Evaluation Metric.} 
We use $\mathtt{Accuracy}$ as the evaluation metric for the above described retrieval task. Here, $\mathtt{Accuracy}$ is defined as the proportion of instances in which the model under evaluation correctly retrieves $\mathbf{A^+}$ over $\mathbf{A^-}$ across the dataset. A higher accuracy score indicates the model is more faithful in encoding numerical information from the text into embeddings, as whether a candidate answer meets the question depends entirely on the numeric values in the text.

\noindent \textbf{Embedding Models.} 
We evaluate a broad range of text embedding models. This includes transformer encoder-based models, such as RoBERTa~\cite{Liu2019RoBERTaAR}, finance-embedding\footnote{\url{https://huggingface.co/FinLang/finance-embeddings-investopedia}}, SimCSE-BERT-unsup~\cite{gao2021simcse}, MiniLM-L6\footnote{\url{https://huggingface.co/sentence-transformers/all-MiniLM-L6-v2}}, and MPNet~\cite{Song2020MPNetMA}, as well as LLM-based models, including gte-Qwen2-7B-instruct~\cite{li2023towards}, Linq-Embed-Mistral~\cite{LinqAIResearch2024}, Qwen3-Embedding-8B~\cite{qwen3embedding}, SFR-Embedding-Mistral~\cite{SFRAIResearch2024}, e5-mistral-7b-instruct~\cite{Wang2023ImprovingTE}, and NV-Embed-v2~\cite{lee2024nv}. Additionally, we consider two proprietary embedding models, Fin-E5~\cite{Tang2025FinMTEBFM} and text-embed-3-small~\cite{openai2024embed}.

\subsection{Main Results}
We report the performance of all evaluated embedding models on the $\mathtt{EmbedNum}$-$\mathtt{1K}$ retrieval task in Table \ref{tab:mainexperiment}. Our main findings are summarized below. 

\begin{table*}[htbp]
\centering
% \begin{tabularx}{\textwidth}{@{}>{\raggedright\noexpandarg}p{4cm} l *{4}{S[table-format=2.1]} *{4}{S[table-format=2.1]}@{}}
\begin{tabularx}{\textwidth}{@{}>{\raggedright}p{4cm} l *{4}{S[table-format=2.1]} *{4}{S[table-format=2.1]}@{}}
\toprule
% \textbf{Model Name} & \multicolumn{1}{c}{\textbf{Orig}} & \multicolumn{4}{c}{\textbf{(a) Integers }} & \multicolumn{4}{c}{\textbf{(b) Decimals}} \\
% \cmidrule(lr){3-6} \cmidrule(lr){7-10}
%  & &  {(a1)} &   {(a2)} &   {(a3)} &   {(a4)} &  {(b1)} &  {(b2)} &  {(b3)} &  {(b4)} \\

\multirow{2}{*}{\textbf{Model Name}} & \multirow{2}{*}{\textbf{Orig}} & \multicolumn{4}{c}{\textbf{(a) Integers}} & \multicolumn{4}{c}{\textbf{(b) Decimals}} \\
\cmidrule(lr){3-6} \cmidrule(lr){7-10}
 & & {(a1)} & {(a2)} & {(a3)} & {(a4)} & {(b1)} & {(b2)} & {(b3)} & {(b4)} \\
 
\midrule
\multicolumn{10}{@{}l}{\textbf{Encoder-Based Models}} \\
\midrule

RoBERTa & 50.8&	50.2	&50.7&	49.5	&49.7	&48.9&	51.2&	49.9	&\underline{50.6} \\
finance-embedding & \underline{53.4}	&\underline{53.4}	&51.2	&\underline{51.7}	&\underline{50.4}&	51.4&	50.1&	51.2&	49.9\\
SimCSE-BERT-unsup & 51.3 &	51.1&	\underline{51.9}&	51.0&	\underline{50.4}&	51.0&	50.7&	50.5	&50.2\\
MiniLM-L6&50.6&	50.2	&51.1	&49.0	&\underline{50.4}&	51.0	&49.8	&50.3&	50.1\\
MPNet& 51.5&53.0 &	49.9 &	50.6&	50.3&	\underline{51.7}	&\underline{51.9}	&\underline{52.9}&	49.7\\

% RoBERTa &\colorbox{red!30}{50.8} & 
% \colorbox{red!30}{50.2} & 
% \colorbox{red!30}{50.7} & 
% \colorbox{red!20}{49.5} & 
% \colorbox{red!20}{49.7} & 
% \colorbox{red!20}{48.9} & 
% \colorbox{red!30}{51.2} & 
% \colorbox{red!20}{49.9} & 
% \colorbox{red!30}{50.6} \\

% finance-embedding & \colorbox{red!30}{53.4} & 
% \colorbox{red!30}{53.4} & 
% \colorbox{red!30}{51.2} & 
% \colorbox{red!30}{51.7} & 
% \colorbox{red!30}{50.4} & 
% \colorbox{red!30}{51.4} & 
% \colorbox{red!30}{50.1} & 
% \colorbox{red!30}{51.2} & 
% \colorbox{red!20}{49.9} \\

% SimCSE-BERT-unsup &\colorbox{red!30}{51.3} & 
% \colorbox{red!30}{51.1} & 
% \colorbox{red!30}{51.9} & 
% \colorbox{red!30}{51.0} & 
% \colorbox{red!30}{50.4} & 
% \colorbox{red!30}{51.0} & 
% \colorbox{red!30}{50.7} & 
% \colorbox{red!30}{50.5} & 
% \colorbox{red!30}{50.2} \\

% MiniLM-L6 & \colorbox{red!30}{50.6} & 
% \colorbox{red!30}{50.2} & 
% \colorbox{red!30}{51.1} & 
% \colorbox{red!20}{49.0} & 
% \colorbox{red!30}{50.4} & 
% \colorbox{red!30}{51.0} & 
% \colorbox{red!20}{49.8} & 
% \colorbox{red!30}{50.3} & 
% \colorbox{red!30}{50.1} \\

% MPNet & \colorbox{red!30}{51.5} & 
% \colorbox{red!30}{53.0} & 
% \colorbox{red!20}{49.9} & 
% \colorbox{red!30}{50.6} & 
% \colorbox{red!30}{50.3} & 
% \colorbox{red!30}{51.7} & 
% \colorbox{red!30}{51.9} & 
% \colorbox{red!30}{52.9} & 
% \colorbox{red!20}{49.7} \\

\midrule

\multicolumn{10}{@{}l}{\textbf{LLM-Based Models}} \\

\midrule

gte-Qwen2-7B-instruct & 51.4& 58.1& 55.2&50.8&	51.3&50.0&51.1&\underline{52.9}&51.8\\
Linq-Embed-Mistral&51.4	&58.0& 52.8& 50.0&48.4&50.7&50.5&50.4&49.0\\
Qwen3-Embedding-8B &\textbf{\underline{53.7}}& \textbf{\underline{62.5}}& \textbf{\underline{55.1}}&\textbf{\underline{53.9}}&\textbf{\underline{52.3}}&51.5&\underline{52.9}&50.9&\underline{52.2}\\
SFR-Embedding-Mistral&52.1& 59.0& 52.4&50.1	&49.2&51.8&	50.6&51.4&50.9\\
e5-mistral-7b-instruct& 51.5& 59.8& 51.3&50.2&	49.2&52.5&51.3&49.7	&50.6\\
NV-Embed-v2&52.1& 62.0& \textbf{\underline{55.1}}& 50.8&	49.1&\underline{53.5}&	52.2&51.2&51.2\\

\midrule
\multicolumn{10}{@{}l}{\textbf{Commercial Models}} \\
\midrule
 Fin-E5  & 51.7 & \underline{58.6} & 52.2 & 48.9 & 48.4 & 51.1 & 48.7 & 49.1 & 49.9 \\
text-embedding-3-small & \underline{52.7} & 50.8 & \underline{54.0} & \underline{51.1} & \underline{50.6} & \textbf{\underline{54.2}} & \textbf{\underline{53.8}} & \textbf{\underline{54.5}} & \textbf{\underline{54.9}} \\
\midrule
% Average performance & 51.9 & 55.9 & 52.5 & 50.6 & 50.0 & 51.5 & 51.1 & 51.1 & 50.8 \\

\textbf{Average performance} &
\cellcolor{lightBlue!45!pureWhite}51.9 &    % 偏浅蓝
\cellcolor{lightRed!50!pureWhite}55.9 &     % 偏浅红
\cellcolor{lightBlue!35!pureWhite}52.5 &    % 浅蓝
\cellcolor{lightBlue!65!pureWhite}50.6 &    % 更浅蓝
\cellcolor{lightBlue!80!pureWhite}50.0 &    % 最浅蓝
\cellcolor{lightBlue!50!pureWhite}51.5 &    % 浅蓝
\cellcolor{lightBlue!55!pureWhite}51.1 &    % 浅蓝
\cellcolor{lightBlue!55!pureWhite}51.1 &    % 浅蓝
\cellcolor{lightBlue!60!pureWhite}50.8 \\

% Fin-E5 & 
% \colorbox{red!30}{51.7} & 
% \colorbox{red!40}{58.6} & 
% \colorbox{red!30}{52.2} & 
% \colorbox{red!20}{48.9} & 
% \colorbox{red!20}{48.4} & 
% \colorbox{red!30}{51.1} & 
% \colorbox{red!20}{48.7} & 
% \colorbox{red!20}{49.1} & 
% \colorbox{red!20}{49.9} \\ 
% text-embedding-3-small	&\colorbox{red!30}{52.7}	&\colorbox{red!30}{50.8} & 	\colorbox{red!30}{54.0}	 & \colorbox{red!30}{51.1}	 & \colorbox{red!30}{50.6}	 & \colorbox{red!30}{54.2} & 	\colorbox{red!30}{53.8}	 & \colorbox{red!30}{54.5}	 & \colorbox{red!30}{54.9} \\
% \midrule
% Avg	&\colorbox{red!30}{51.9} & \colorbox{red!40}{55.9} & 	\colorbox{red!30}{52.5} & \colorbox{red!30}{50.6} & 	\colorbox{red!30}{50.0} 	& \colorbox{red!30}{51.5} & 	\colorbox{red!30}{51.1} 	& \colorbox{red!30}{51.1} & 	\colorbox{red!30}{50.8}  \\

\midrule
\bottomrule

\multirow{2}{*}{\textbf{Model Name}} & \multicolumn{3}{c}{\textbf{(c) Scaled Down}} & \multicolumn{3}{c}{\textbf{(d) Scaled Up}} & \multicolumn{3}{c}{\textbf{(e) Others}} \\
\cmidrule(lr){2-4} \cmidrule(lr){5-7} \cmidrule(lr){8-10}

% \textbf{Model Name} & \multicolumn{3}{c}{\textbf{(c) Scaled Down}} & \multicolumn{3}{c}{\textbf{(d) Scaled Up}} & \multicolumn{3}{c}{\textbf{(e) Others}} \\

% \cmidrule(lr){2-4} \cmidrule(lr){5-7} \cmidrule(lr){8-10}

  % & {10} & {100} & {1000} & {10} & {100} & {1000} & {percent} & {\%} & {C} \\
  & {(c1)}   & {(c2)}   & {(c3)}  & {(d1)}  & {(d2)}  & {(d3)} & {(e1)}  & {(e2)}  & {(e3)} \\
\midrule
\multicolumn{10}{@{}l}{\textbf{Encoder-Based Models}} \\
\midrule
RoBERTa& 50.4&	50.2&	51.6&52.4&	50.8&	47.3&	52.5&	51.9&	50.7 \\
finance-embedding & \underline{54.0}&	49.9& 53.9&	\underline{52.7}&	\underline{53.6}&	51.1&	\underline{54.2}&	\underline{54.8}	&51.0 \\
SimCSE-BERT-unsup & 52.1&	\underline{53.3}&	52.2&	50.7&	52.7&	51.0&	51.8&	52.4&	50.9 \\
MiniLM-L6  & 51.4	&50.1&	49.2&	52.5&	50.4&	51.1&	50.7&	51.4&	50.7 \\
MPNet & 52.5&	52.6	&\underline{55.2}&	50.6&	51.4&	\underline{53.2}&	{53.7}	&53.8	&\underline{51.9} \\
% RoBERTa & 
% \colorbox{red!30}{50.4} & 
% \colorbox{red!30}{50.2} & 
% \colorbox{red!30}{51.6} & 
% \colorbox{red!30}{52.4} & 
% \colorbox{red!30}{50.8} & 
% \colorbox{red!20}{47.3} & 
% \colorbox{red!30}{52.5} & 
% \colorbox{red!30}{51.9} & 
% \colorbox{red!30}{50.7} \\

% finance-embedding & 
% \colorbox{red!30}{54.0} & 
% \colorbox{red!20}{49.9} & 
% \colorbox{red!30}{53.9} & 
% \colorbox{red!30}{52.7} & 
% \colorbox{red!30}{53.6} & 
% \colorbox{red!30}{51.1} & 
% \colorbox{red!30}{54.2} & 
% \colorbox{red!30}{54.8} & 
% \colorbox{red!30}{51.0} \\

% SimCSE-BERT-unsup & 
% \colorbox{red!30}{52.1} & 
% \colorbox{red!30}{53.3} & 
% \colorbox{red!30}{52.2} & 
% \colorbox{red!30}{50.7} & 
% \colorbox{red!30}{52.7} & 
% \colorbox{red!30}{51.0} & 
% \colorbox{red!30}{51.8} & 
% \colorbox{red!30}{52.4} & 
% \colorbox{red!30}{50.9} \\

% MiniLM-L6 & 
% \colorbox{red!30}{51.4} & 
% \colorbox{red!30}{50.1} & 
% \colorbox{red!20}{49.2} & 
% \colorbox{red!30}{52.5} & 
% \colorbox{red!30}{50.4} & 
% \colorbox{red!30}{51.1} & 
% \colorbox{red!30}{50.7} & 
% \colorbox{red!30}{51.4} & 
% \colorbox{red!30}{50.7} \\

% MPNet & 
% \colorbox{red!30}{52.5} & 
% \colorbox{red!30}{52.6} & 
% \colorbox{red!40}{55.2} & 
% \colorbox{red!30}{50.6} & 
% \colorbox{red!30}{51.4} & 
% \colorbox{red!30}{53.2} & 
% \colorbox{red!30}{53.7} & 
% \colorbox{red!30}{53.8} & 
% \colorbox{red!30}{51.9} \\ 

\midrule
\multicolumn{10}{@{}l}{\textbf{LLM-Based Models}} \\
\midrule

gte-Qwen2-7B-instruct & 65.3 & 63.9 & 57.8 & 58.3 & 51.0 & 52.9 & 61.9 & 57.4 & 51.3 \\

Linq-Embed-Mistral & 67.3 & 58.2 & 54.8 & 55.3 & 49.8 & 46.9 & 64.4 & 64.9 & 49.7 \\

Qwen3-Embedding-8B & 69.2 & 67.9 & 64.7 & 61.0 &  \textbf{\underline{55.8}} &  \textbf{\underline{54.7}} &  \textbf{\underline{71.0}} &  \textbf{\underline{69.9}} & \textbf{\underline{53.3}} \\

SFR-Embedding-Mistral & 67.9 & 61.5 & 57.6 & 53.1 & 49.9 & 48.6 & 65.5 & 64.7 & 48.6 \\

e5-mistral-7b-instruct & 66.5 & 61.2 & 57.4 & 53.5 & 50.8 & 49.3 & 67.2 & 64.8 & 49.3 \\

NV-Embed-v2 & \textbf{\underline{74.2}} & \textbf{\underline{70.6} }&  \textbf{\underline{65.1}} &  \textbf{\underline{61.9}} & 54.7 & 51.7 & 69.3 & 66.6 & 49.7 \\

\midrule
\multicolumn{10}{@{}l}{\textbf{Commercial Models}} \\
\midrule

Fin-E5 & 62.8 & 58.4 & 54.2 & 54.0 & \underline{52.2} & 48.6 & \underline{66.6} & \underline{67.9} & 49.3 \\
text-embed-3-small & \underline{67.6} & \underline{59.6} & \underline{58.0} & \underline{57.9} & 51.3 & \underline{50.2} & 59.2 & 55.0 & \underline{50.5} \\
\midrule
% Average performance & 61.6 & 58.3 & 56.3 & 54.9 & 51.9 & 50.5 & 60.6 & 59.7 & 50.5 \\
\textbf{Average performance} &
\cellcolor{lightRed!100!pureWhite}61.6 &    % 最大值 → 浅红
\cellcolor{lightRed!70!pureWhite}58.3 &     % 偏浅红
\cellcolor{lightRed!60!pureWhite}56.3 &     % 浅红
\cellcolor{lightBlue!20!pureWhite}54.9 &    % 接近白色
\cellcolor{lightBlue!45!pureWhite}51.9 &    % 浅蓝
\cellcolor{lightBlue!70!pureWhite}50.5 &    % 较浅蓝
\cellcolor{lightRed!90!pureWhite}60.6 &     % 偏浅红
\cellcolor{lightRed!80!pureWhite}59.7 &     % 偏浅红
\cellcolor{lightBlue!70!pureWhite}50.5 \\
\bottomrule

% Fin-E5 & 
% \colorbox{red!50}{62.8} & 
% \colorbox{red!40}{58.4} & 
% \colorbox{red!30}{54.2} & 
% \colorbox{red!30}{54.0} & 
% \colorbox{red!30}{52.2} & 
% \colorbox{red!20}{48.6} & 
% \colorbox{red!60}{66.6} & 
% \colorbox{red!60}{67.9} & 
% \colorbox{red!20}{49.3} \\ 
% text-embed-3-small  & \colorbox{red!60}{67.6}&\colorbox{red!40}{59.6} &\colorbox{red!40}{58.0}& \colorbox{red!40}{57.9}&\colorbox{red!30}{51.3} &\colorbox{red!30}{50.2} &\colorbox{red!40}{59.2} &\colorbox{red!40}{55.0}&\colorbox{red!30}{50.5}    \\

% \midrule
% Avg	& \colorbox{red!50}{61.6}& 	\colorbox{red!40}{58.3} 	&\colorbox{red!40}{56.3}&	\colorbox{red!30}{54.9} &\colorbox{red!30}{51.9}	&\colorbox{red!30}{50.5} 	&\colorbox{red!50}{60.6} &	\colorbox{red!40}{59.7} &	\colorbox{red!30}{50.5} \\
% \bottomrule

\end{tabularx}
\caption{Retrieval accuracy (\%) on the original $\mathtt{EmbedNum}$-$\mathtt{1K}$ data and its 17 variants across 13 evaluated text embedding models. For each numeric format, the highest accuracy is highlighted in \textbf{bold}, and the highest accuracy within each model category is \underline{underlined}. A \textcolor{blue}{blue}-white-\textcolor{red}{red} color gradient is overlaid on the ``Average performance'' row, with warmer colors indicating higher accuracy and cooler colors indicating lower accuracy across numeric formats.}
\label{tab:mainexperiment}
\end{table*}

\noindent \textbf{Overall, embedding models struggle to capture numerical details in text.} 
The retrieval accuracy, averaged over the 13 embedding models under evaluation, is merely 0.54, slightly above random guessing (0.5).

We believe this is mainly because current embedding model training practices typically focus on capturing semantic similarities or differences at the sentence or paragraph level, with fine-grained but potentially important details, such as numbers as part of the text, often overlooked. 
This aligns with \citet{liu2024beyond}, who find that embedding models generally fall short in distinguishing nuanced differences between texts. Our finding further confirms this limitation in the specific context of number-related nuances. 
We suggest that future work may consider not treating numbers on par with words, and should instead explore dedicated design considerations for numerical content to improve embedding models' ability to model numbers in text. 

\noindent \textbf{LLM-based embedding models show a clear advantage over encoder-based models.} 
We observe that LLM-based models achieve, on average, about 5 percentage points higher accuracy than encoder-based ones (0.56 vs. 0.51). 

We speculate that this advantage is largely due to the pre-existing superior natural language understanding capability of an LLM prior to its adaptation for embedding purposes. 
However, through what mechanisms this superiority meaningfully translates into tangible gains in embedding quality is still an open question and certainly merits further investigation. 

\noindent \textbf{Embedding models interpret 8\% and 0.08 differently.} 
Our experimental results reveal that embedding models are quite sensitive to how numbers are represented in text, with accuracy varying by up to 12 percentage points depending on numeric format, doing best on 0.x numbers (e.g., 0.5, 0.8) with an accuracy of 0.62 and struggling most with 4-digit integers (e.g., 1234, 5678) at a random guessing level. 

For this reason, we strongly recommend that future efforts in developing new number modeling strategies carefully account for numeric formats, given their critical role in how models interpret numbers as part of text.

\noindent \textbf{Improved literacy does not necessarily translate into tangible gains in numeracy.} 
Among the evaluated models, Fin-E5 is a financially adapted version of e5-mistral-7b-instruct, further fine-tuned on financial data. 
Since the evaluation data $\mathtt{EmbedNum}$-$\mathtt{1K}$ is specific to finance, it allows us to examine whether domain-specific fine-tuning (i.e., improved financial literacy) helps models understand numbers (i.e., enhanced numeracy). 
By comparing their performance, we find no evidence that Fin-E5 clearly wins over the base model. 

Complementary to the observation by \citet{thawani2021numeracy} that better numeracy leads to better literacy, our finding underscores the unique difficulty of numeracy compared to literacy and again highlights the need for specialized designs directly targeting numeracy.

\noindent \textbf{OOV (out-of-vocabulary) numbers challenge embedding models.} 
Analyzing the results based on number formatting, we find that embedding models consistently struggle with decimal numbers (e.g., 8.67), large integers (e.g., 3035), and comma-separated numbers (e.g., 3,035). 
A closer look reveals that numbers in these formats are often out-of-vocabulary (OOV), meaning they are more likely to be split into multiple smaller units (tokens) during tokenization. 
For example, in a comma-separated number, the comma often marks a boundary, leading to a number like ``1,100'' being broken into ``1'', ``,'', and ``100''. 

This could be problematic because splitting a number can disrupt key information, such as its magnitude, which is essential for many downstream tasks. 
Thus, we believe addressing this OOV issue has great potential to improve the usefulness of current embedding models in number-intensive real-world applications.

\noindent \textbf{Number precision affects model performance, reflecting human cognition.} 
The consistently higher accuracies achieved on numbers such as 0.8, 0.02, 0.007, and 0.0006 prompt us to think about the potential reasons behind this pattern. 
Our in-depth analysis reveals an interesting phenomenon: model performance is highly correlated with the number of significant figures (s.f.) in the numbers the model processes. 
More specifically, as shown in Figure~\ref{fig:signumber}, as the number of significant figures increases, the mean accuracy over the 13 embedding models for the corresponding numbers declines. 

This observation suggests that, just as humans experience higher cognitive load and are more prone to errors when processing numbers with many digits, embedding models also struggle with long, high-precision numbers. 
This represents an interesting research avenue in language model-brain alignment \citep{yu2024predicting}. 

\begin{figure}[htbp] 
\centering 
\includegraphics[width=1\columnwidth]{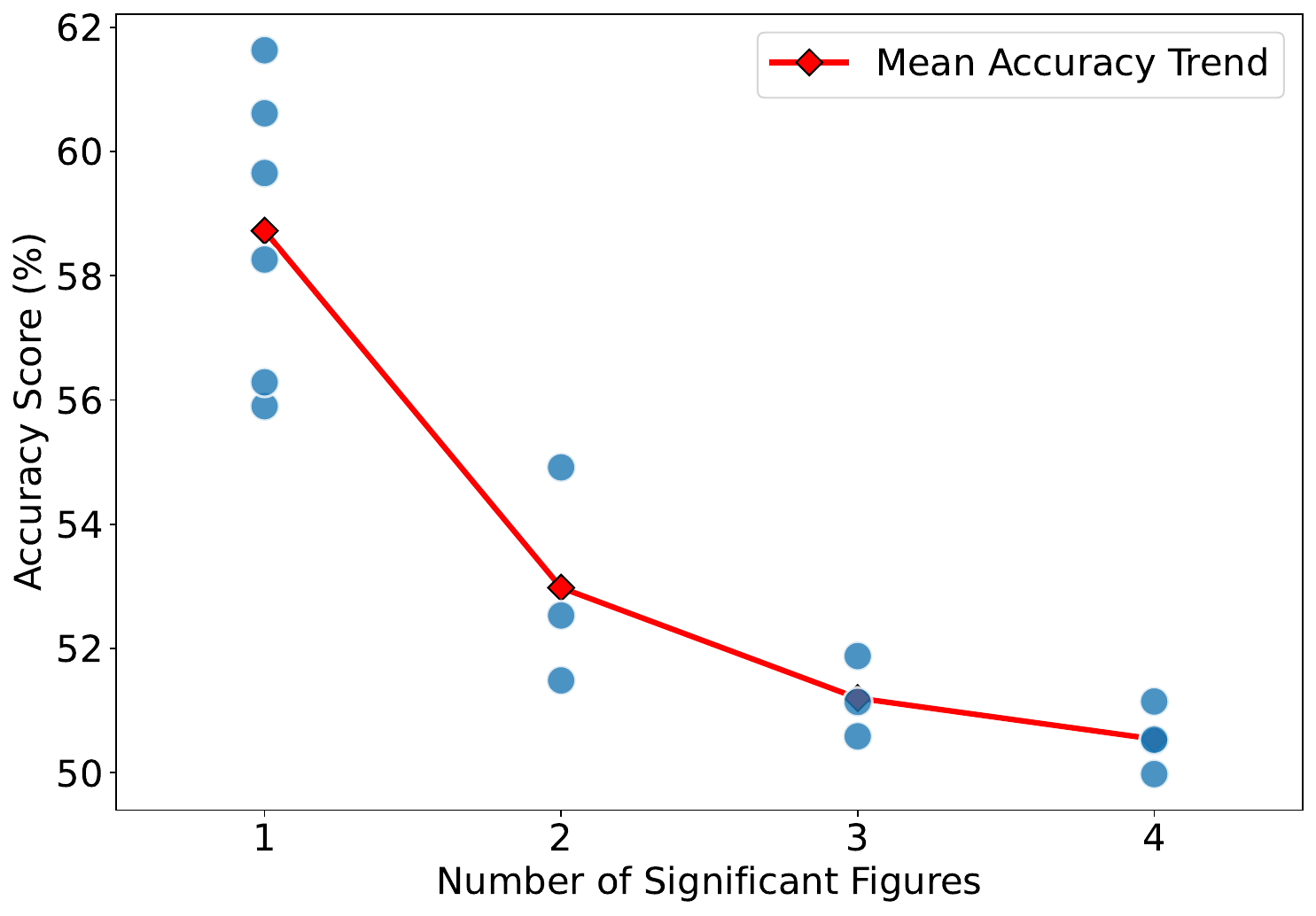} 
\caption{Each blue dot represents the mean performance across the 13 embedding models for a specific numeric format. Each numeric format is categorized by the number of significant figures in the numbers expressed under that format, shown on the X-axis.} 
\label{fig:signumber} 
\end{figure}

\section{Additional Analyses}
\subsection{Evidence of Performance Asymmetry Across Numeric Conditions}
In the main experiments, we report accuracy calculated over all instances in the dataset. 
Here, we take a closer look and examine the model performance separately for two subsets of data, stratified by the numeric condition specified in each question ($\mathbf{Q}$). 
Specifically, one subset includes questions asking for answers above a threshold (greater-than condition), while the other includes questions asking for answers below a threshold (less-than condition). 
For instance, the question ``Which stocks have a P/E ratio over 200?'' represents a greater-than condition, while ``Which stocks have a P/E ratio under 200?'' represents a less-than condition. 

Figure~\ref{fig:lessgreat} compares model performance on these two categories of data. 
It demonstrates clear evidence that embedding models systematically favor answering greater-than questions over less-than questions, with markedly higher accuracy. 
This asymmetry suggests that \textbf{embedding models may not be equally reliable across all numeric reasoning scenarios}, and this may have significant implications in clinical and financial decision-making, where errors in certain types of numeric comparisons can have especially serious consequences. 
This observation serves as a reminder to apply embedding models with care in such settings.
\begin{figure}[htbp] 
\centering 
\includegraphics[width=1\columnwidth]{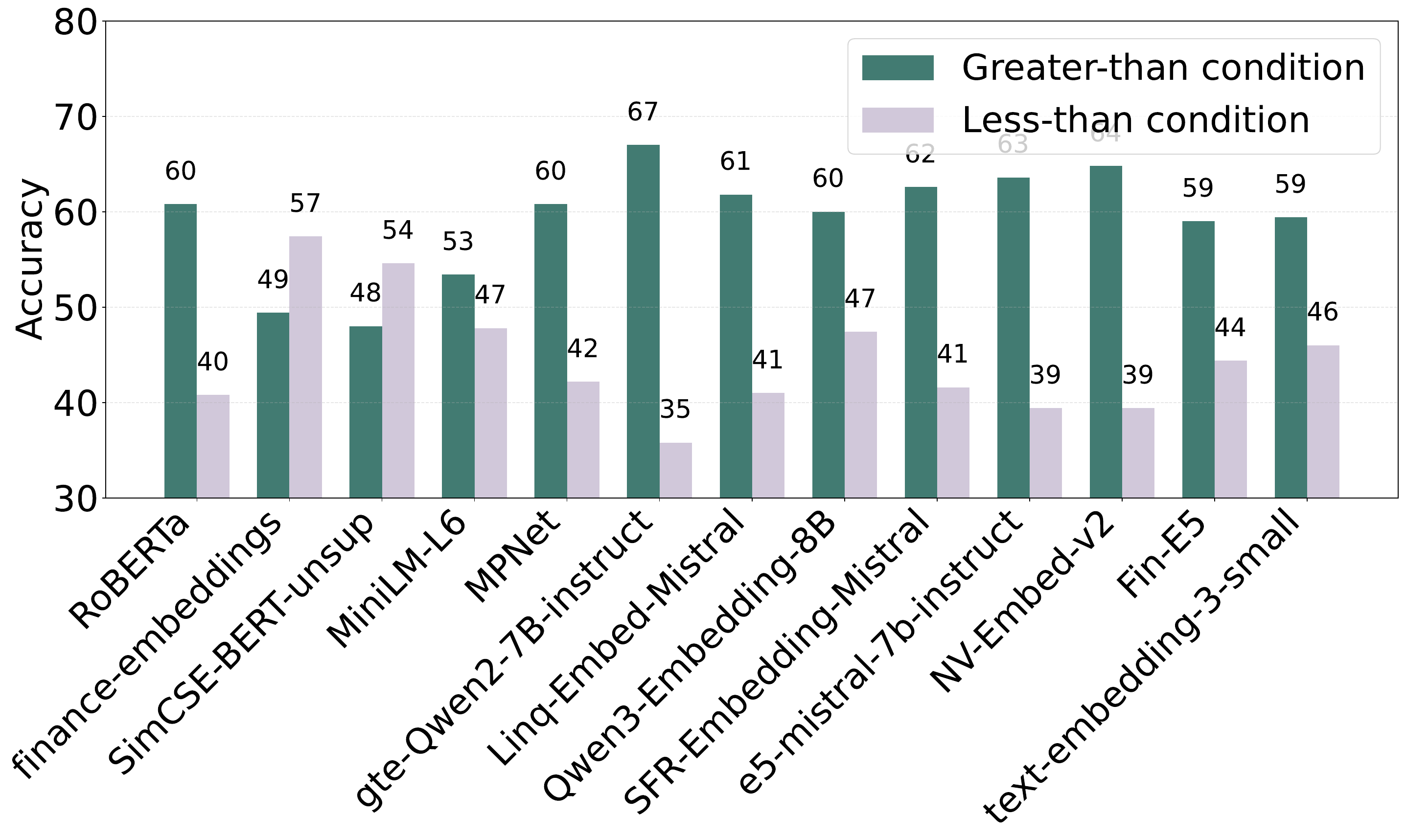} 
\caption{Model performance by numeric condition (greater-than vs. less-than).} 
\label{fig:lessgreat} 
\end{figure}

\subsection{``Twenty-Four'' versus ``24''}
We investigate whether rewriting numbers in their written form (e.g., 24 as ``twenty-four'') alleviates the cognitive effort required by embedding models to interpret numerical content. 
Our evaluation covers integers with up to four digits (e.g., 6789) and their corresponding written forms (e.g., six thousand seven hundred and eighty-nine). 

Figure~\ref{fig:wordformat} compares model performance on these two forms of numbers, where we observe that most models achieve higher accuracy on the written-form numbers, albeit with a minor advantage. 
We speculate that this marginal gain is largely because written numbers are often composed of in-vocabulary word tokens, allowing key information to be better preserved. 

However, even with written numbers, overall accuracy remains low, suggesting that \textbf{representing numerical content is an inherent limitation of embedding models rather than solely an OOV issue}. 
This calls for more reliable approaches to overcome the challenge, beyond simply expanding the model's vocabulary to cover numbers that were previously absent.

\begin{figure}[htbp] 
\centering 
\includegraphics[width=0.98\columnwidth]{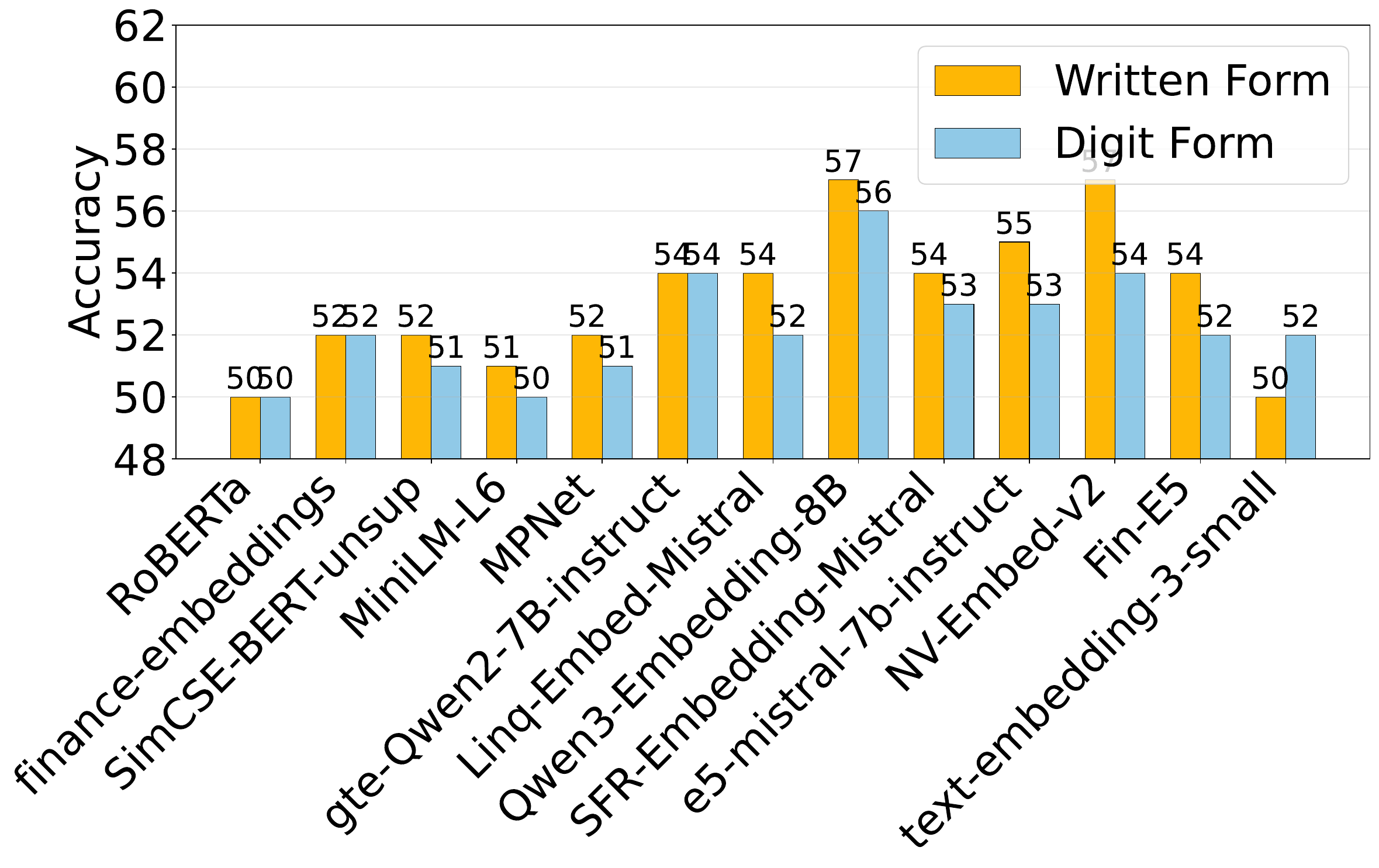} 
\caption{Model performance by numeric representation (digit form vs. written form).} 
\label{fig:wordformat} 
\end{figure}

\subsection{Does Seeing Numbers More Often Help Models Handle Them Better?}

Previous studies suggest that the $\ell^2$-norm of a token's embedding reflects how often that token appears in the training data \citep{li-etal-2020-sentence, yu-etal-2022-rare}. 
Specifically, tokens with smaller $\ell^2$-norms tend to appear more frequently in the training data, whereas rarer tokens generally have larger embedding norms. 
This inspires us to examine whether higher frequency exposure to numbers can translate into tangible numeracy gains. 
Concretely, we stratify the evaluation data based on the $\ell^2$-norms of the uncontextualized number embeddings (in the model's embedding lookup table), forming a high-frequency group from the bottom half of $\ell^2$-norms and a low-frequency group from the top half of $\ell^2$-norms. 
Please refer to Figure~\ref{fig:financeembedding} for an illustration, which suggests that very small numbers (e.g., 1, 2) and very large numbers (e.g., 1800, 2000) appear more frequently in training, while mid-range values are relatively underrepresented and associated with larger embedding norms.

\begin{figure}[htbp] 
\centering 
\includegraphics[width=1\columnwidth]{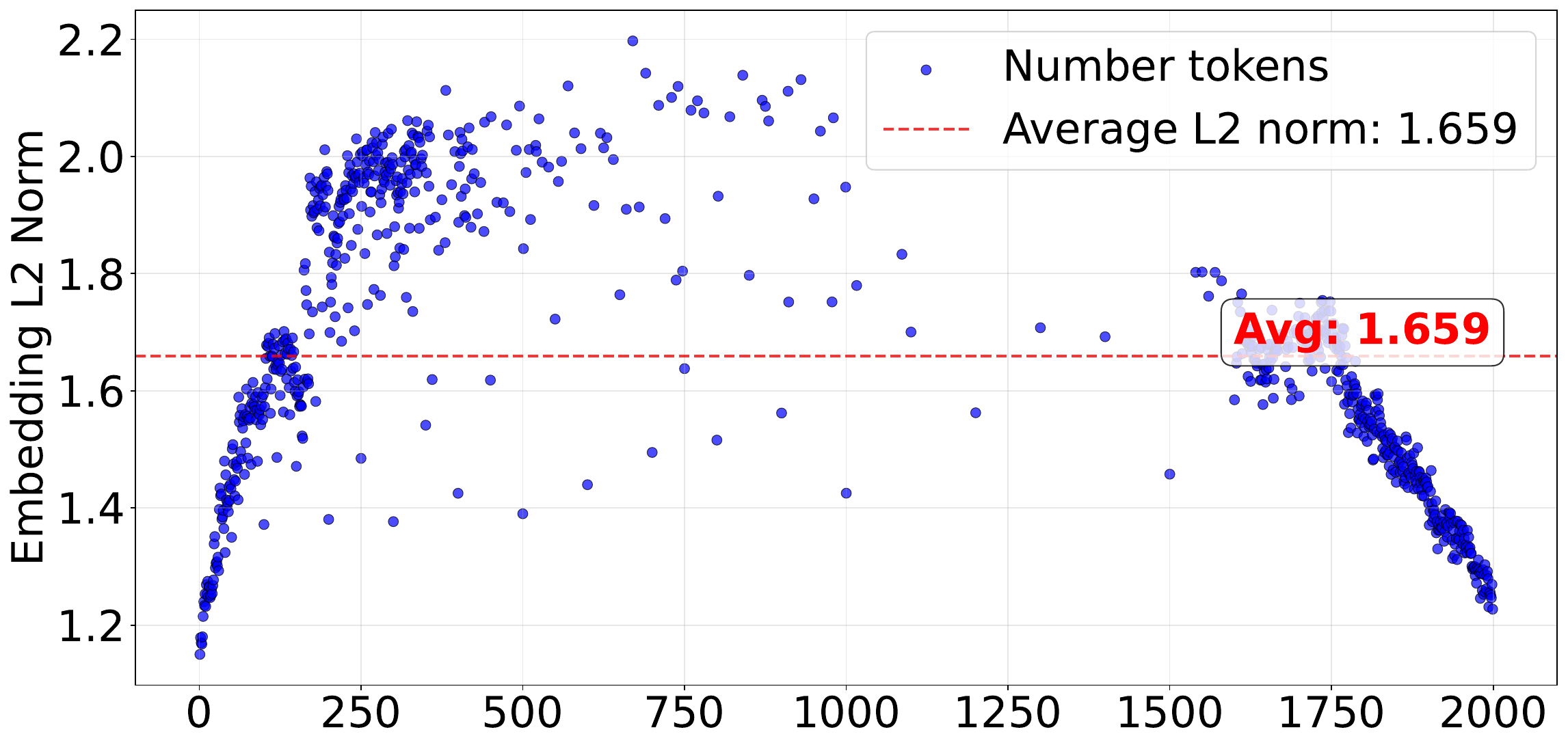} 
\caption{A visualization of the $\ell^2$-norm distribution of number embeddings from the finance-embedding token lookup table.} 
\label{fig:financeembedding} 
\end{figure}
We present model performance across the two frequency-based groups in Figure~\ref{fig:toptail}. Surprisingly, the results do not indicate a clear, consistent performance advantage for high-frequency numbers over low-frequency ones. 
This suggests that \textbf{simply making pretraining data more number-intensive is unlikely to yield meaningful gains in numeracy and might be less cost-efficient}; instead, more targeted approaches for modeling numbers are needed.
\begin{figure}[htbp] 
\centering 
\includegraphics[width=1\columnwidth]{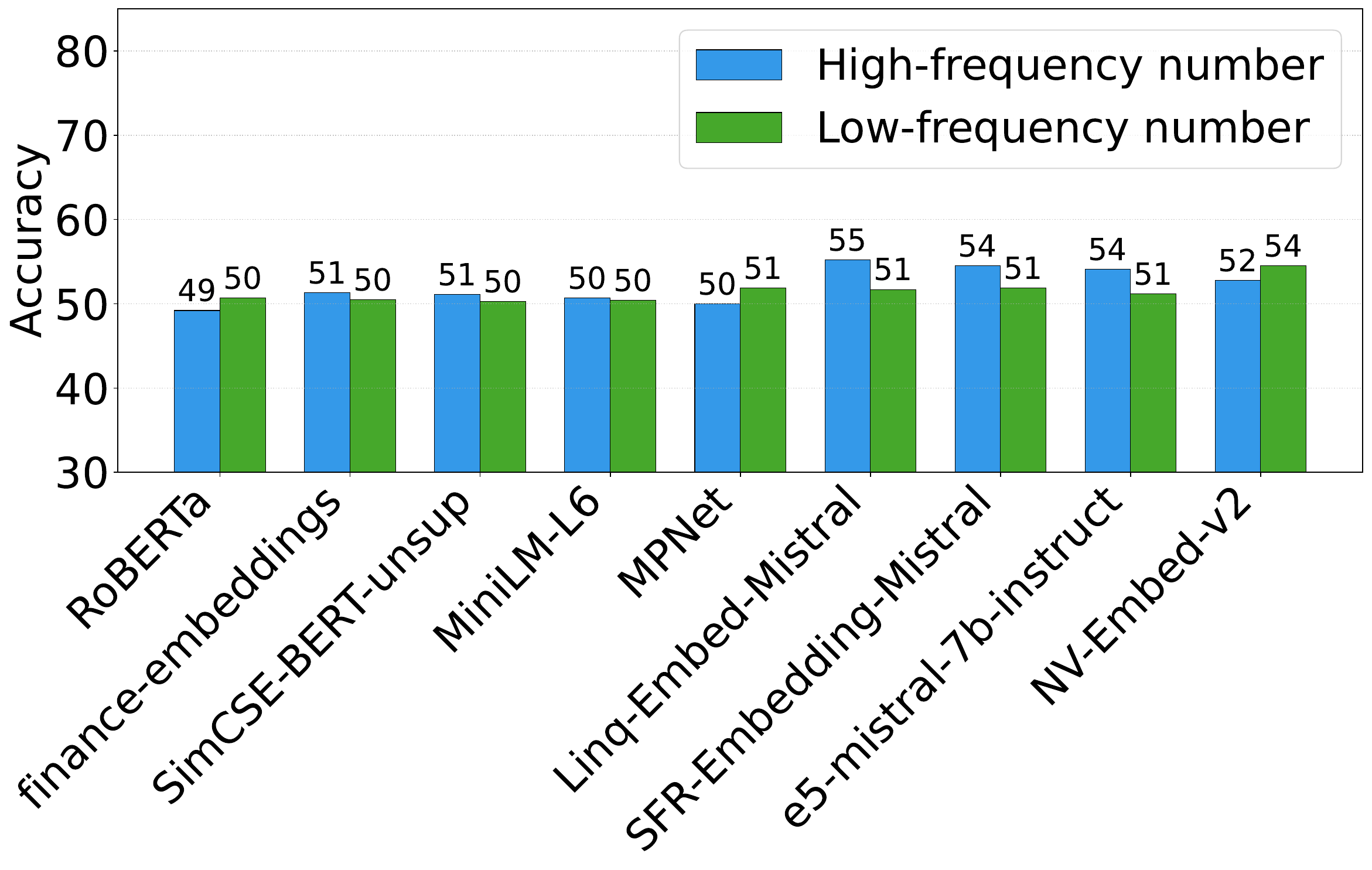} 
\caption{Model performance by number frequency (high- vs. low-frequency numbers).} 
\label{fig:toptail} 
\end{figure}

\subsection{Does Context Hide the Numbers? A Trade-Off in Embeddings}
Previous studies highlight a granularity dilemma in embeddings, where there is often a trade-off between capturing the overall semantic meaning of a text and preserving its fine-grained details \citep{xu2025dense}. 
Inspired by this, we investigate whether a similar trade-off occurs in our numeracy-focused setting. For example, does the ``P/E ratio'' context in ``Which stocks have a P/E ratio above 200?'' weaken the numerical signal for 200 compared to a context-free question like ``Which number is above 200?'' 

To study this, we synthesize 1000 sentences of the form ``The freezing and pledging party's largest shareholder, John Smith, holds a \{X\} shareholding,'' where \{X\} takes integer values from 1 to 1000. 
Using Qwen3-Embedding-8B \citep{qwen3embedding}, we encode these sentences and compute the cosine similarity between embeddings of adjacent numbers (e.g., \{X\} and \{X+1\}). 
We present the similarity score distribution in Figure \ref{fig:unif}, comparing it to a context-free baseline where only the numbers appear, i.e., computing the cosine similarity between embeddings of \{X\} and \{X+1\}. 
We observe that in the context-rich setting, similarity scores collapse near one, indicating that \textbf{most of the embedding's capacity is taken up by the semantic context, weakening the representation of fine-grained numerical details}.

\begin{figure}[htbp] 
\centering 
\includegraphics[width=1\columnwidth]{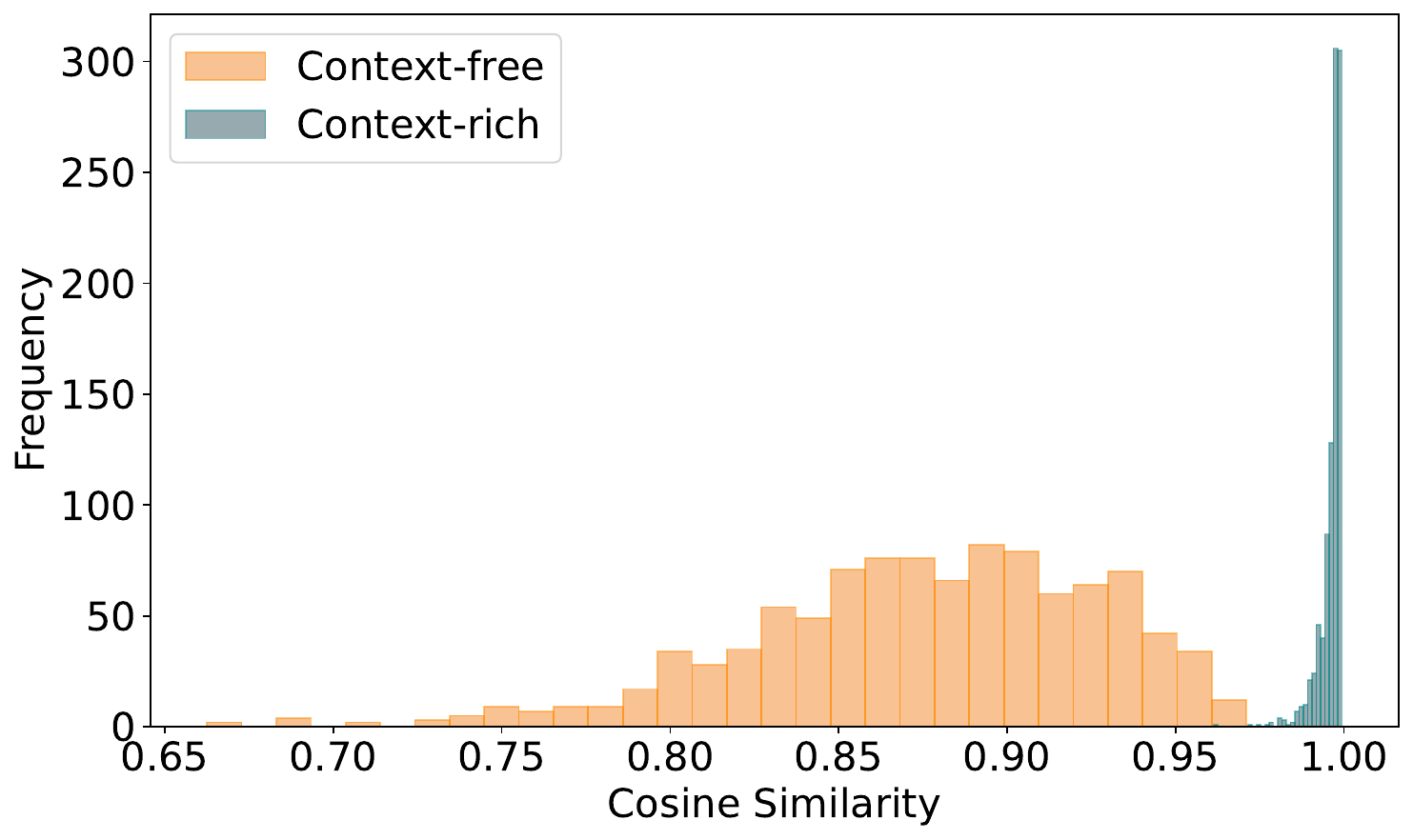} 
\caption{Distribution of embedding similarities by context condition (context-rich vs. context-free).} 
\label{fig:unif} 
\end{figure}

To further examine whether this effect accounts for the overall low performance on $\mathtt{EmbedNum}$-$\mathtt{1K}$, as reported in the main experiments, we create a context-reduced version of the original dataset, e.g., adapting ``Which stocks have a P/E ratio above 200?'' to ``Which number is above 200?'' with answers ``220'' and ``180''. 

As shown in Figure~\ref{fig:context_vs_withoutcontext}, we find that models generally perform better in the context-reduced setting, implying that \textbf{additional context appears to attenuate the fine-grained numerical signals in embeddings}. 
This confirms the granularity dilemma \citep{xu2025dense} and, in turn, suggests that expanding embedding capacity or expressivity may constitute an essential step toward enhancing the numeracy of embeddings.

\begin{figure}[htbp] 
\centering 
\includegraphics[width=1\columnwidth]{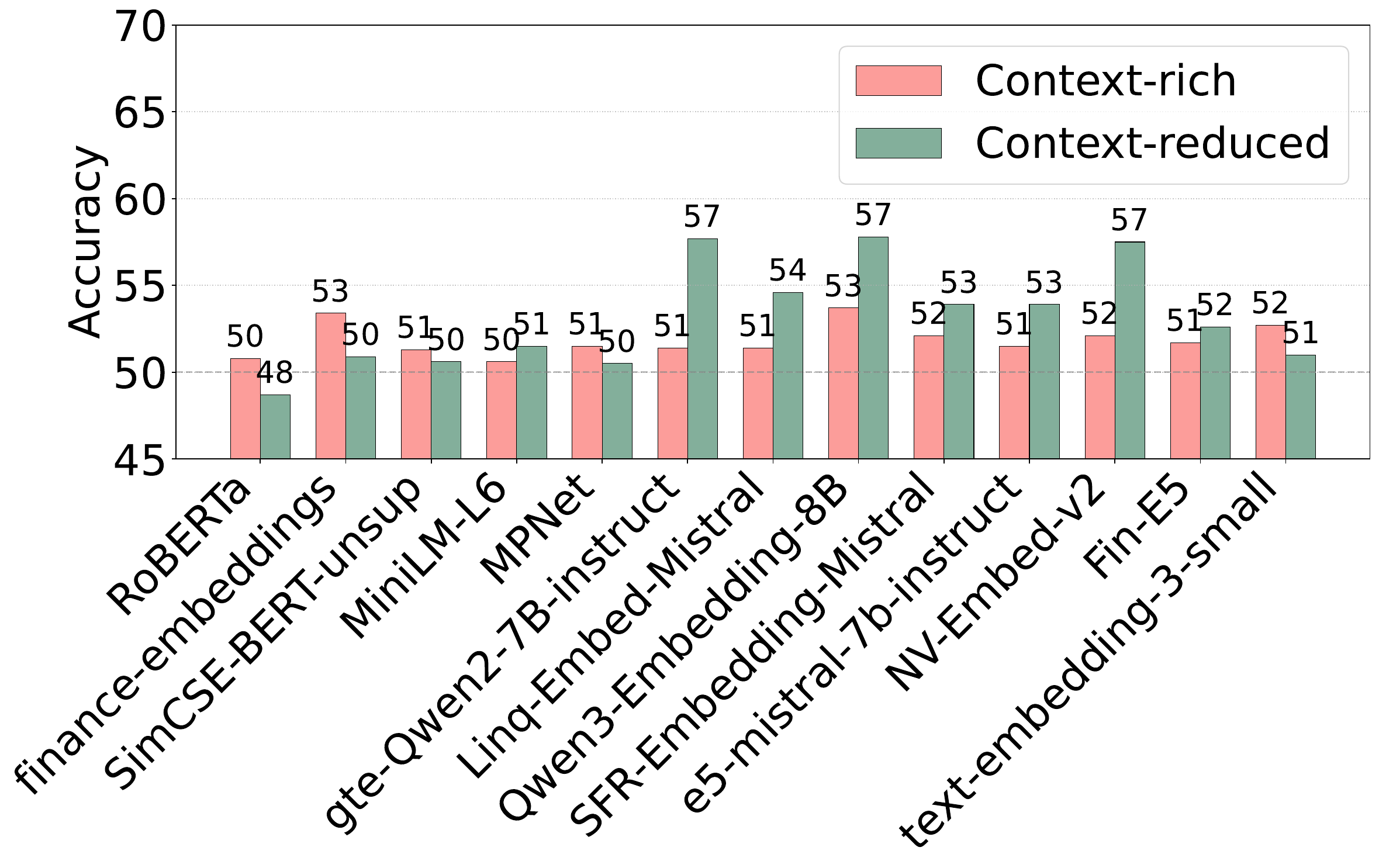} 
\caption{Model performance by context condition (context-rich vs. context-reduced).} 
\label{fig:context_vs_withoutcontext} 
\end{figure}

\subsection{Can Probing Predict Downstream Numeracy Performance?}
Probing has long been used to assess numeracy in word embeddings \citep{naik2019exploring, wallace2019nlp}, such as GloVe \citep{pennington2014glove} or word2vec \citep{mikolov2013efficient}, by training a simple regressor to decode numerical values from word embeddings (e.g., mapping ``seven'' to 7). 
We extend this idea to text embeddings, where numbers appear as part of sentences or paragraphs (e.g., ``Stock ABC has a P/E of 220.'') rather than as individual tokens (e.g., ``220''). 
Specifically, we train a linear regressor on 70\% of $\mathtt{EmbedNum}$-$\mathtt{1K}$ answer embeddings to predict the numerical value of the number contained in the answer, and evaluate it on the remaining 30\%, reporting  Adjusted   $\text{R}^2$ scores (Table~\ref{tab: adjusted}). 

Surprisingly, encoder-based models consistently yield higher  Adjusted  $\text{R}^2$ scores than LLM-based models, meaning that numerical values are easier to probe from encoder embeddings. 

This observation contrasts with our main finding in the earlier $\mathtt{EmbedNum}$-$\mathtt{1K}$ retrieval task, where LLM-based models outperform encoder-based ones. This suggests that \textbf{embeddings from which numerical values are more easily decoded do not necessarily imply more faithful preservation of meaningful numeracy for downstream tasks}. 
This questions the reliability of probing tests as an evaluation metric in context-rich settings and highlights the need for new metrics to assess embedding numeracy in such environments.

\begin{table}[htbp]
\centering
\setlength\tabcolsep{4pt}
\resizebox{0.476\textwidth}{!}{%
\begin{tabular}{lcc}
\toprule
\textbf{Model Name} & \textbf{Embedding Dimension} & \textbf{Adjusted R\textsuperscript{2}} \\
\toprule
\multicolumn{3}{l}{\textbf{Encoder-Based Models}} \\
\midrule
RoBERTa & 768 & \cellcolor{red!45!white}1.25 \\
finance-embeddings & 768 & \cellcolor{red!55!white}1.99 \\
SimCSE-BERT-unsup & 768 & \cellcolor{red!40!white}1.13 \\
MiniLM-L6 & 384 & \cellcolor{red!60!white}4.98 \\
MPNet & 768 & \cellcolor{red!42!white}1.15 \\
\midrule
\multicolumn{3}{l}{\textbf{LLM-Based Models}} \\
\midrule
gte-Qwen2-7B-instruct & 3584 & \cellcolor{blue!10!white}1.03 \\
Linq-Embed-Mistral & 4096 & \cellcolor{blue!35!white}1.01 \\
Qwen3-Embedding-8B & 4096 & \cellcolor{blue!25!white}1.02 \\
SFR-Embedding-Mistral & 4096 & \cellcolor{blue!35!white}1.01 \\
e5-mistral-7b-instruct & 4096 & \cellcolor{blue!35!white}1.01 \\
NV-Embed-v2 & 4096 & \cellcolor{blue!35!white}1.01 \\
\midrule
\multicolumn{3}{l}{\textbf{Commercial Models}} \\
\midrule
Fin-E5 & 4096 & \cellcolor{blue!35!white}1.01 \\
text-embedding-3-small & 1536 & \cellcolor{blue!10!white}1.03 \\
\bottomrule
\end{tabular}%
}
\caption{Results of probing numerical values from text embeddings. Higher Adjusted $\text{R}^2$ values indicate a more accurate prediction of numerical values. A \textcolor{blue}{blue}-white-\textcolor{red}{red} color gradient is overlaid on the Adjusted $\text{R}^2$ column, with warmer colors representing higher values and cooler colors representing lower values.}
\label{tab: adjusted}
\end{table}

\section{Conclusion}
In this study, we examine numeracy in text embeddings. 
Our work addresses an important evaluation gap, where fine-grained numerical nuances in text are often overlooked in existing embedding model benchmarking. 
Our experimental results reveal a significant limitation of modern text embedding models in preserving meaningful numerical details within their embeddings. 
This limitation is observed in both encoder-based and LLM-based models. 
Through further analyses, we gain deeper insights into embedding numeracy and provide potential guidance for improving embedding models' ability to manage numerical content. 
We hope this work can shed light on future research to strengthen embedding-based NLP systems in number-intensive application scenarios.

\section{Limitations} 
Our work also has limitations that can be improved in future research.
First, we focus primarily on the comparison of number magnitudes, which is a common type of numerical reasoning. Other forms of numerical reasoning, such as arithmetic operations, ratio and percentage calculations, or time comparisons, are not explored. Future work could extend our evaluation to include these more complex reasoning tasks. 
Second, our study relies on synthetic datasets to evaluate embedding numeracy. While synthetic data enables controlled experiments and clear analysis, it may not reflect the full complexity of real-world contexts, where numbers interact with more complex linguistic and contextual information. Future research could identify representative real-world applications and develop related benchmarks, and investigate embedding numeracy across diverse domains, such as healthcare, scientific texts, and other areas where accurate numerical understanding is essential. 
Finally, we focus this work primarily on sentence-level embeddings and do not investigate embeddings of larger text units, such as paragraphs or full documents. Future studies could build on our evaluation framework to examine numeracy in embeddings that encode larger contexts.

\iftaclpubformat

\appendix
\section{Including acknowledgments}
Acknowledgments appear immediately before the references.  Do not number this
section.\footnote{In \LaTeX, one can use {\tt {\textbackslash}section*} instead
of {\tt {\textbackslash}section}.} If you found the reviewers' or Action
Editor's comments helpful; consider acknowledging them.
\else
\fi

\bibliography{embedding}

\begin{thebibliography}{36}
\expandafter\ifx\csname natexlab\endcsname\relax\def\natexlab#1{#1}\fi

\bibitem[{DeepSeek-AI(2024)}]{deepseekai2024deepseekv3technicalreport}
DeepSeek-AI. 2024.
\newblock \href {http://arxiv.org/abs/2412.19437} {Deepseek-v3 technical report}.

\bibitem[{Duan et~al.(2021)Duan, Yang, and Tam}]{duan2021learning}
Hanyu Duan, Yi~Yang, and Kar~Yan Tam. 2021.
\newblock Learning numeracy: A simple yet effective number embedding approach using knowledge graph.
\newblock In \emph{Findings of the Association for Computational Linguistics: EMNLP 2021}, pages 2597--2602.

\bibitem[{Enevoldsen et~al.(2025)Enevoldsen, Chung, Kerboua, Kardos, Mathur, Stap, Gala, Siblini, Krzemi'nski, Winata, Sturua, Utpala, Ciancone, Schaeffer, Sequeira, Misra, Dhakal, Rystr{\o}m, Solomatin, cCaugatan, Kundu, Bernstorff, Xiao, Sukhlecha, Pahwa, Poswiata, Kranthikiran, Ashraf, Auras, Pluster, Harries, Magne, Mohr, Hendriksen, Zhu, Gisserot-Boukhlef, Aarsen, Kostkan, Wojtasik, Lee, vSuppa, Zhang, Rocca, Hamdy, Michail, Yang, Faysse, Vatolin, Thakur, Dey, Vasani, Chitale, Tedeschi, Tai, Snegirev, Gunther, Xia, Shi, L{\`u}, Clive, Krishnakumar, Maksimova, Wehrli, Tikhonova, Panchal, Abramov, Ostendorff, Liu, Clematide, Miranda, Fenogenova, Song, Safi, Li, Borghini, Cassano, Su, Lin, Yen, Hansen, Hooker, Xiao, Adlakha, Weller, Reddy, and Muennighoff}]{Enevoldsen2025MMTEBMM}
Kenneth~C. Enevoldsen, Isaac Chung, Imene Kerboua, M{\'a}rton Kardos, Ashwin Mathur, David Stap, Jay Gala, Wissam Siblini, Dominik Krzemi'nski, Genta~Indra Winata, Saba Sturua, Saiteja Utpala, Mathieu Ciancone, Marion Schaeffer, Gabriel Sequeira, Diganta Misra, Shreeya Dhakal, Jonathan Rystr{\o}m, Roman~Sergeevich Solomatin, Omer~Veysel cCaugatan, Akash Kundu, Martin Bernstorff, Shitao Xiao, Akshita Sukhlecha, Bhavish Pahwa, Rafal Poswiata, G~Kranthikiran, Shawon Ashraf, Daniel Auras, Bjorn Pluster, Jan~Philipp Harries, Loic Magne, Isabelle Mohr, Mariya Hendriksen, Dawei Zhu, Hippolyte Gisserot-Boukhlef, Tom Aarsen, Jan Kostkan, Konrad Wojtasik, Taemin Lee, Marek vSuppa, Crystina Zhang, Roberta Rocca, Mohammed Hamdy, Andrianos Michail, John Yang, Manuel Faysse, Aleksei Vatolin, Nandan Thakur, Manan Dey, Dipam Vasani, Pranjal~A. Chitale, Simone Tedeschi, Nguyen Tai, Artem Snegirev, Michael Gunther, Mengzhou Xia, Weijia Shi, Xing~Han L{\`u}, Jordan Clive, Gayatri Krishnakumar, Anna Maksimova, Silvan Wehrli,
  Maria Tikhonova, Henil Panchal, Aleksandr Abramov, Malte Ostendorff, Zheng Liu, Simon Clematide, Lester James~Validad Miranda, Alena Fenogenova, Guangyu Song, Ruqiya~Bin Safi, Wen-Ding Li, Alessia Borghini, Federico Cassano, Hongjin Su, Jimmy Lin, Howard Yen, Lasse Hansen, Sara Hooker, Chenghao Xiao, Vaibhav Adlakha, Orion Weller, Siva Reddy, and Niklas Muennighoff. 2025.
\newblock \href {https://api.semanticscholar.org/CorpusID:276449901} {Mmteb: Massive multilingual text embedding benchmark}.
\newblock \emph{ArXiv}, abs/2502.13595.

\bibitem[{Fan et~al.(2024)Fan, Ding, Ning, Wang, Li, Yin, Chua, and Li}]{fan2024survey}
Wenqi Fan, Yujuan Ding, Liangbo Ning, Shijie Wang, Hengyun Li, Dawei Yin, Tat-Seng Chua, and Qing Li. 2024.
\newblock A survey on rag meeting llms: Towards retrieval-augmented large language models.
\newblock In \emph{Proceedings of the 30th ACM SIGKDD conference on knowledge discovery and data mining}, pages 6491--6501.

\bibitem[{Gao et~al.(2021)Gao, Yao, and Chen}]{gao2021simcse}
Tianyu Gao, Xingcheng Yao, and Danqi Chen. 2021.
\newblock Simcse: Simple contrastive learning of sentence embeddings.
\newblock In \emph{Proceedings of the 2021 Conference on Empirical Methods in Natural Language Processing}, pages 6894--6910. Association for Computational Linguistics.

\bibitem[{Jiang et~al.(2020)Jiang, Nian, Guo, Chu, Zhao, Shen, and Tu}]{jiang2020learning}
Chengyue Jiang, Zhonglin Nian, Kaihao Guo, Shanbo Chu, Yinggong Zhao, Libin Shen, and Kewei Tu. 2020.
\newblock Learning numeral embedding.
\newblock In \emph{Findings of the Association for Computational Linguistics: EMNLP 2020}, pages 2586--2599.

\bibitem[{Kim et~al.(2024)Kim, Lee, Kwon, Gu, Kim, Cho, yong Sohn, and Choi}]{LinqAIResearch2024}
Junseong Kim, Seolhwa Lee, Jihoon Kwon, Sangmo Gu, Yejin Kim, Minkyung Cho, Jy~yong Sohn, and Chanyeol Choi. 2024.
\newblock \href {https://getlinq.com/blog/linq-embed-mistral/} {Linq-embed-mistral:elevating text retrieval with improved gpt data through task-specific control and quality refinement}.
\newblock Linq AI Research Blog.

\bibitem[{Lee et~al.(2024)Lee, Roy, Xu, Raiman, Shoeybi, Catanzaro, and Ping}]{lee2024nv}
Chankyu Lee, Rajarshi Roy, Mengyao Xu, Jonathan Raiman, Mohammad Shoeybi, Bryan Catanzaro, and Wei Ping. 2024.
\newblock Nv-embed: Improved techniques for training llms as generalist embedding models.
\newblock \emph{arXiv preprint arXiv:2405.17428}.

\bibitem[{Li et~al.(2020)Li, Zhou, He, Wang, Yang, and Li}]{li-etal-2020-sentence}
Bohan Li, Hao Zhou, Junxian He, Mingxuan Wang, Yiming Yang, and Lei Li. 2020.
\newblock \href {https://doi.org/10.18653/v1/2020.emnlp-main.733} {On the sentence embeddings from pre-trained language models}.
\newblock In \emph{Proceedings of the 2020 Conference on Empirical Methods in Natural Language Processing (EMNLP)}, pages 9119--9130, Online. Association for Computational Linguistics.

\bibitem[{Li et~al.(2023)Li, Zhang, Zhang, Long, Xie, and Zhang}]{li2023towards}
Zehan Li, Xin Zhang, Yanzhao Zhang, Dingkun Long, Pengjun Xie, and Meishan Zhang. 2023.
\newblock Towards general text embeddings with multi-stage contrastive learning.
\newblock In \emph{Proceedings of the 2023 Conference on Empirical Methods in Natural Language Processing}. Association for Computational Linguistics.

\bibitem[{Liu et~al.(2024)Liu, Yang, and Tam}]{liu2024beyond}
Jiaxin Liu, Yi~Yang, and Kar~Yan Tam. 2024.
\newblock Beyond surface similarity: Detecting subtle semantic shifts in financial narratives.
\newblock In \emph{Findings of the Association for Computational Linguistics: NAACL 2024}, pages 2641--2652.

\bibitem[{Liu et~al.(2019)Liu, Ott, Goyal, Du, Joshi, Chen, Levy, Lewis, Zettlemoyer, and Stoyanov}]{Liu2019RoBERTaAR}
Yinhan Liu, Myle Ott, Naman Goyal, Jingfei Du, Mandar Joshi, Danqi Chen, Omer Levy, Mike Lewis, Luke Zettlemoyer, and Veselin Stoyanov. 2019.
\newblock \href {https://api.semanticscholar.org/CorpusID:198953378} {Roberta: A robustly optimized bert pretraining approach}.
\newblock \emph{ArXiv}, abs/1907.11692.

\bibitem[{Meng et~al.(2024)Meng, Liu, Joty, Xiong, Zhou, and Yavuz}]{SFRAIResearch2024}
Rui Meng, Ye~Liu, Shafiq~Rayhan Joty, Caiming Xiong, Yingbo Zhou, and Semih Yavuz. 2024.
\newblock \href {https://www.salesforce.com/blog/sfr-embedding/} {Sfr-embedding-mistral:enhance text retrieval with transfer learning}.
\newblock Salesforce AI Research Blog.

\bibitem[{Mikolov et~al.(2013)Mikolov, Chen, Corrado, and Dean}]{mikolov2013efficient}
Tomas Mikolov, Kai Chen, Greg Corrado, and Jeffrey Dean. 2013.
\newblock Efficient estimation of word representations in vector space.
\newblock \emph{arXiv preprint arXiv:1301.3781}.

\bibitem[{Muennighoff(2022)}]{muennighoff2022sgpt}
Niklas Muennighoff. 2022.
\newblock Sgpt: Gpt sentence embeddings for semantic search.
\newblock \emph{arXiv preprint arXiv:2202.08904}.

\bibitem[{Muennighoff et~al.(2023)Muennighoff, Tazi, Magne, and Reimers}]{muennighoff-etal-2023-mteb}
Niklas Muennighoff, Nouamane Tazi, Loic Magne, and Nils Reimers. 2023.
\newblock \href {https://doi.org/10.18653/v1/2023.eacl-main.148} {{MTEB}: Massive text embedding benchmark}.
\newblock In \emph{Proceedings of the 17th Conference of the European Chapter of the Association for Computational Linguistics}, pages 2014--2037, Dubrovnik, Croatia. Association for Computational Linguistics.

\bibitem[{Naik et~al.(2019)Naik, Ravichander, Rose, and Hovy}]{naik2019exploring}
Aakanksha Naik, Abhilasha Ravichander, Carolyn Rose, and Eduard Hovy. 2019.
\newblock Exploring numeracy in word embeddings.
\newblock In \emph{Proceedings of the 57th Annual Meeting of the Association for Computational Linguistics}, pages 3374--3380.

\bibitem[{{OpenAI}(2025)}]{openai2024embed}
{OpenAI}. 2025.
\newblock Openai (august 25 version).
\newblock \url{https://api.openai.com/v1/embeddings}.

\bibitem[{Pennington et~al.(2014)Pennington, Socher, and Manning}]{pennington2014glove}
Jeffrey Pennington, Richard Socher, and Christopher~D Manning. 2014.
\newblock Glove: Global vectors for word representation.
\newblock In \emph{Proceedings of the 2014 conference on empirical methods in natural language processing (EMNLP)}, pages 1532--1543.

\bibitem[{Sivakumar and Moosavi(2023)}]{Sivakumar2023FERMATAA}
Jasivan Sivakumar and Nafise~Sadat Moosavi. 2023.
\newblock \href {https://api.semanticscholar.org/CorpusID:258959201} {Fermat: An alternative to accuracy for numerical reasoning}.
\newblock \emph{ArXiv}, abs/2305.17491.

\bibitem[{Sivakumar and Moosavi(2025)}]{sivakumar2025leverage}
Jasivan~Alex Sivakumar and Nafise~Sadat Moosavi. 2025.
\newblock How to leverage digit embeddings to represent numbers?
\newblock In \emph{Proceedings of the 31st International Conference on Computational Linguistics}, pages 7685--7697.

\bibitem[{Song et~al.(2020)Song, Tan, Qin, Lu, and Liu}]{Song2020MPNetMA}
Kaitao Song, Xu~Tan, Tao Qin, Jianfeng Lu, and Tie-Yan Liu. 2020.
\newblock \href {https://api.semanticscholar.org/CorpusID:215827489} {Mpnet: Masked and permuted pre-training for language understanding}.
\newblock \emph{ArXiv}, abs/2004.09297.

\bibitem[{Sundararaman et~al.(2020)Sundararaman, Si, Subramanian, Wang, Hazarika, and Carin}]{sundararaman2020methods}
Dhanasekar Sundararaman, Shijing Si, Vivek Subramanian, Guoyin Wang, Devamanyu Hazarika, and Lawrence Carin. 2020.
\newblock Methods for numeracy-preserving word embeddings.
\newblock In \emph{Proceedings of the 2020 Conference on Empirical Methods in Natural Language Processing (EMNLP)}, pages 4742--4753.

\bibitem[{Tang and Yang(2025)}]{Tang2025FinMTEBFM}
Yixuan Tang and Yi~Yang. 2025.
\newblock \href {https://api.semanticscholar.org/CorpusID:276409274} {Finmteb: Finance massive text embedding benchmark}.
\newblock \emph{ArXiv}, abs/2502.10990.

\bibitem[{Thakur et~al.(2021)Thakur, Reimers, Ruckl'e, Srivastava, and Gurevych}]{Thakur2021BEIRAH}
Nandan Thakur, Nils Reimers, Andreas Ruckl'e, Abhishek Srivastava, and Iryna Gurevych. 2021.
\newblock \href {https://api.semanticscholar.org/CorpusID:233296016} {Beir: A heterogenous benchmark for zero-shot evaluation of information retrieval models}.
\newblock \emph{ArXiv}, abs/2104.08663.

\bibitem[{Thawani et~al.(2021)Thawani, Pujara, and Ilievski}]{thawani2021numeracy}
Avijit Thawani, Jay Pujara, and Filip Ilievski. 2021.
\newblock Numeracy enhances the literacy of language models.
\newblock In \emph{Proceedings of the 2021 conference on empirical methods in natural language processing}, pages 6960--6967.

\bibitem[{Wallace et~al.(2019)Wallace, Wang, Li, Singh, and Gardner}]{wallace2019nlp}
Eric Wallace, Yizhong Wang, Sujian Li, Sameer Singh, and Matt Gardner. 2019.
\newblock Do nlp models know numbers? probing numeracy in embeddings.
\newblock In \emph{Proceedings of the 2019 Conference on Empirical Methods in Natural Language Processing and the 9th International Joint Conference on Natural Language Processing (EMNLP-IJCNLP)}, pages 5307--5315.

\bibitem[{Wang et~al.(2023)Wang, Yang, Huang, Yang, Majumder, and Wei}]{Wang2023ImprovingTE}
Liang Wang, Nan Yang, Xiaolong Huang, Linjun Yang, Rangan Majumder, and Furu Wei. 2023.
\newblock \href {https://api.semanticscholar.org/CorpusID:266693831} {Improving text embeddings with large language models}.
\newblock \emph{ArXiv}, abs/2401.00368.

\bibitem[{Wong et~al.(2025)Wong, Ali, Xiong, Shen, Kim, and Agrawal}]{wong2025position}
Lionel Wong, Ayman Ali, Raymond~M Xiong, Zejiang Shen, Yoon Kim, and Monica Agrawal. 2025.
\newblock \href {https://openreview.net/forum?id=LL39y0Tfxb} {Position: Retrieval-augmented systems can be dangerous medical communicators}.
\newblock In \emph{Forty-second International Conference on Machine Learning Position Paper Track}.

\bibitem[{Xu et~al.(2025)Xu, Su, Yu, Li, Meng, and Zhou}]{xu2025dense}
Liyan Xu, Zhenlin Su, Mo~Yu, Jiangnan Li, Fandong Meng, and Jie Zhou. 2025.
\newblock Dense retrievers can fail on simple queries: Revealing the granularity dilemma of embeddings.
\newblock \emph{arXiv preprint arXiv:2506.08592}.

\bibitem[{Yepes et~al.(2024)Yepes, You, Milczek, Laverde, and Li}]{yepes2024financial}
Antonio~Jimeno Yepes, Yao You, Jan Milczek, Sebastian Laverde, and Renyu Li. 2024.
\newblock Financial report chunking for effective retrieval augmented generation.
\newblock \emph{arXiv preprint arXiv:2402.05131}.

\bibitem[{Yu et~al.(2022)Yu, Song, Kim, Lee, Ryu, and Yoon}]{yu-etal-2022-rare}
Sangwon Yu, Jongyoon Song, Heeseung Kim, Seongmin Lee, Woo-Jong Ryu, and Sungroh Yoon. 2022.
\newblock \href {https://doi.org/10.18653/v1/2022.acl-long.3} {Rare tokens degenerate all tokens: Improving neural text generation via adaptive gradient gating for rare token embeddings}.
\newblock In \emph{Proceedings of the 60th Annual Meeting of the Association for Computational Linguistics (Volume 1: Long Papers)}, pages 29--45, Dublin, Ireland. Association for Computational Linguistics.

\bibitem[{Yu et~al.(2024)Yu, Gu, Huang, and Li}]{yu2024predicting}
Shaoyun Yu, Chanyuan Gu, Kexin Huang, and Ping Li. 2024.
\newblock Predicting the next sentence (not word) in large language models: What model-brain alignment tells us about discourse comprehension.
\newblock \emph{Science advances}, 10(21):eadn7744.

\bibitem[{Zhang et~al.(2024)Zhang, Mao, Fan, Mi, Gao, Chen, Lou, and Lin}]{Zhang2024FinSQLML}
Chao Zhang, Yuren Mao, Yijiang Fan, Yu~Mi, Yunjun Gao, Lu~Chen, Dongfang Lou, and Jinshu Lin. 2024.
\newblock \href {https://api.semanticscholar.org/CorpusID:267061057} {Finsql: Model-agnostic llms-based text-to-sql framework for financial analysis}.
\newblock \emph{Companion of the 2024 International Conference on Management of Data}.

\bibitem[{Zhang et~al.(2025{\natexlab{a}})Zhang, Zhang, Zhao, Huang, Hu, and Zhang}]{zhang2025role}
Meishan Zhang, Xin Zhang, Xinping Zhao, Shouzheng Huang, Baotian Hu, and Min Zhang. 2025{\natexlab{a}}.
\newblock On the role of pretrained language models in general-purpose text embeddings: A survey.
\newblock \emph{arXiv preprint arXiv:2507.20783}.

\bibitem[{Zhang et~al.(2025{\natexlab{b}})Zhang, Li, Long, Zhang, Lin, Yang, Xie, Yang, Liu, Lin, Huang, and Zhou}]{qwen3embedding}
Yanzhao Zhang, Mingxin Li, Dingkun Long, Xin Zhang, Huan Lin, Baosong Yang, Pengjun Xie, An~Yang, Dayiheng Liu, Junyang Lin, Fei Huang, and Jingren Zhou. 2025{\natexlab{b}}.
\newblock Qwen3 embedding: Advancing text embedding and reranking through foundation models.
\newblock \emph{arXiv preprint arXiv:2506.05176}.

\end{thebibliography}
\bibliographystyle{acl_natbib}

\iftaclpubformat

\onecolumn

% \begin{figure}[htbp] 
% \centering 
% \includegraphics[width=0.98\columnwidth]{figs/5.2.1.pca_bge_large_en.pdf} 
% \caption{The distribution of similarity and uniformity among numbers} 
% \label{fig:5.2.1} 
% \end{figure}

% \begin{figure}[htbp] 
% \centering 
% \includegraphics[width=0.98\columnwidth]{figs/5.2.2.pca_bge_large_en.pdf} 
% \caption{The distribution of similarity and uniformity among sentences with a finance context} 
% \label{fig:5.2.2} 
% \end{figure}

\appendix
\section{Author/Affiliation Options as set forth by MIT Press}
\label{sec:authorformatting}

Option 1. The author’s address is underneath each name, centered.

\begin{quote}\centering
  \begin{tabular}{c}
    \textbf{First Author} \\
    First Affiliation \\
    First Address 1 \\
    First Address 2 \\
    \texttt{first.email@example.com}
  \end{tabular}
  \ 
  \begin{tabular}{c}
    \textbf{Second Author} \\
    Second Affiliation \\
    Second Address 1 \\
    Second Address 2 \\
    \texttt{second.email@example.com}
  \end{tabular}

  \begin{tabular}{c}
    \textbf{Third Author} \\
    Third Affiliation \\
    Third Address 1 \\
    Third Address 2 \\
    \texttt{third.email@example.com}
  \end{tabular}
\end{quote}

Option 2. Author’s address is linked with superscript characters to its name,
author names are grouped, centered.

\begin{quote}\centering
    \textbf{First Author$^\diamond$} \quad \textbf{Second Author$^\dagger$} \quad
    \textbf{Third Author$^\ddagger$}
    \\ \ \\
    $^\diamond$First Affiliation \\
    First Address 1 \\
    First Address 2 \\
    \texttt{first.email@example.com}
     \\ \ \\
     $^\dagger$Second Affiliation \\
    Second Address 1 \\
    Second Address 2 \\
    \texttt{second.email@example.com}
     \\ \ \\
    $^\ddagger$Third Affiliation \\
    Third Address 1 \\
    Third Address 2 \\
    \texttt{third.email@example.com}
\end{quote}
  
\fi

\clearpage
\appendix
\section{Prompts Used for Data Curation}
\label{appx:dataconstruction}

\vspace{40pt}
% 第一个图片 - 使用 minipage 避免浮动
\noindent
\begin{minipage}{\textwidth}
\centering
\includegraphics[width=1\textwidth]{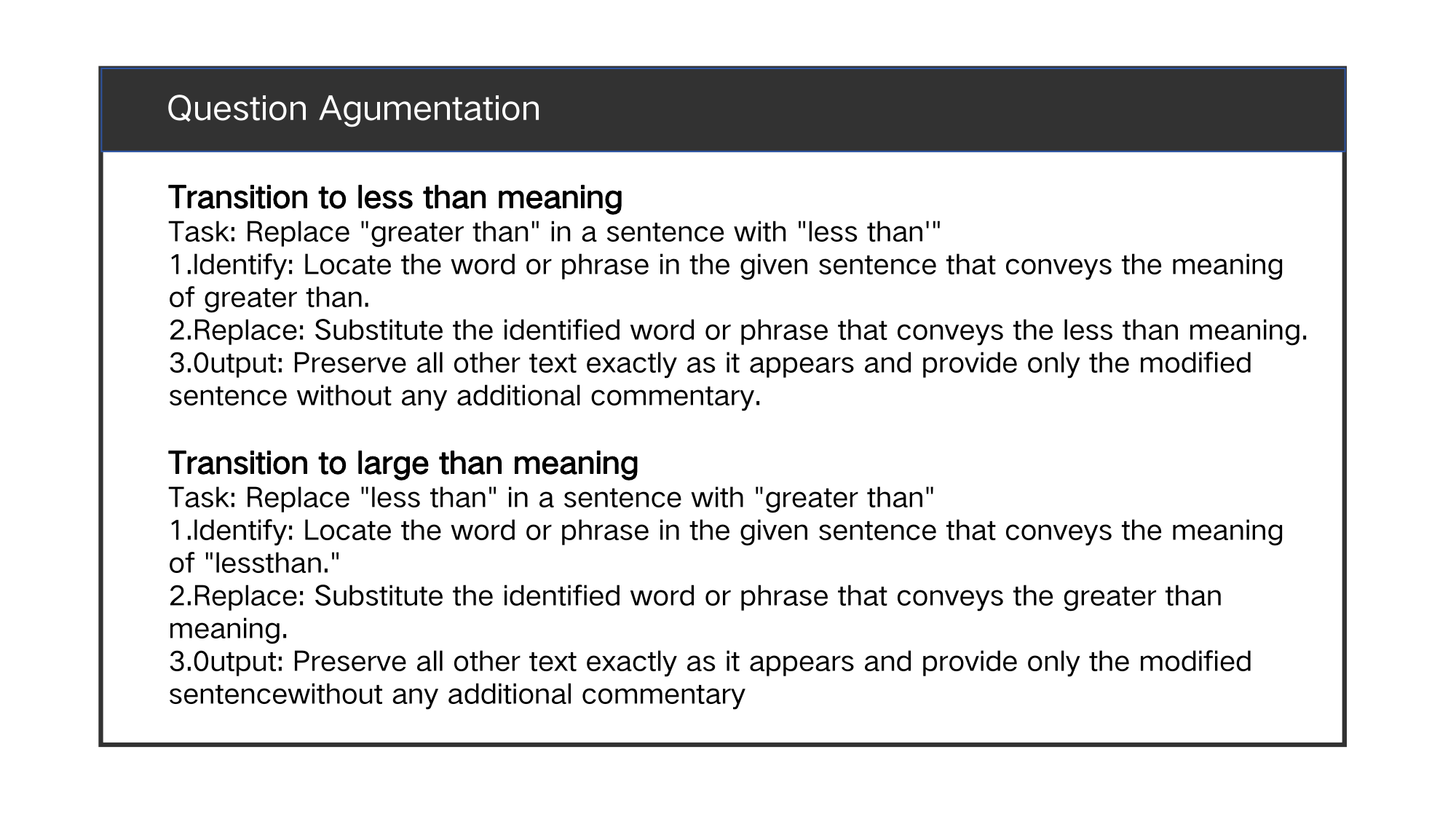}
\par\vspace{10pt}
Figure 9: Prompt used for question augmentation.
\label{fig:semanticprompt}
\end{minipage}

\vspace{50pt}

% 第二个图片
\noindent
\begin{minipage}{\textwidth}
\centering
\includegraphics[width=1\textwidth]{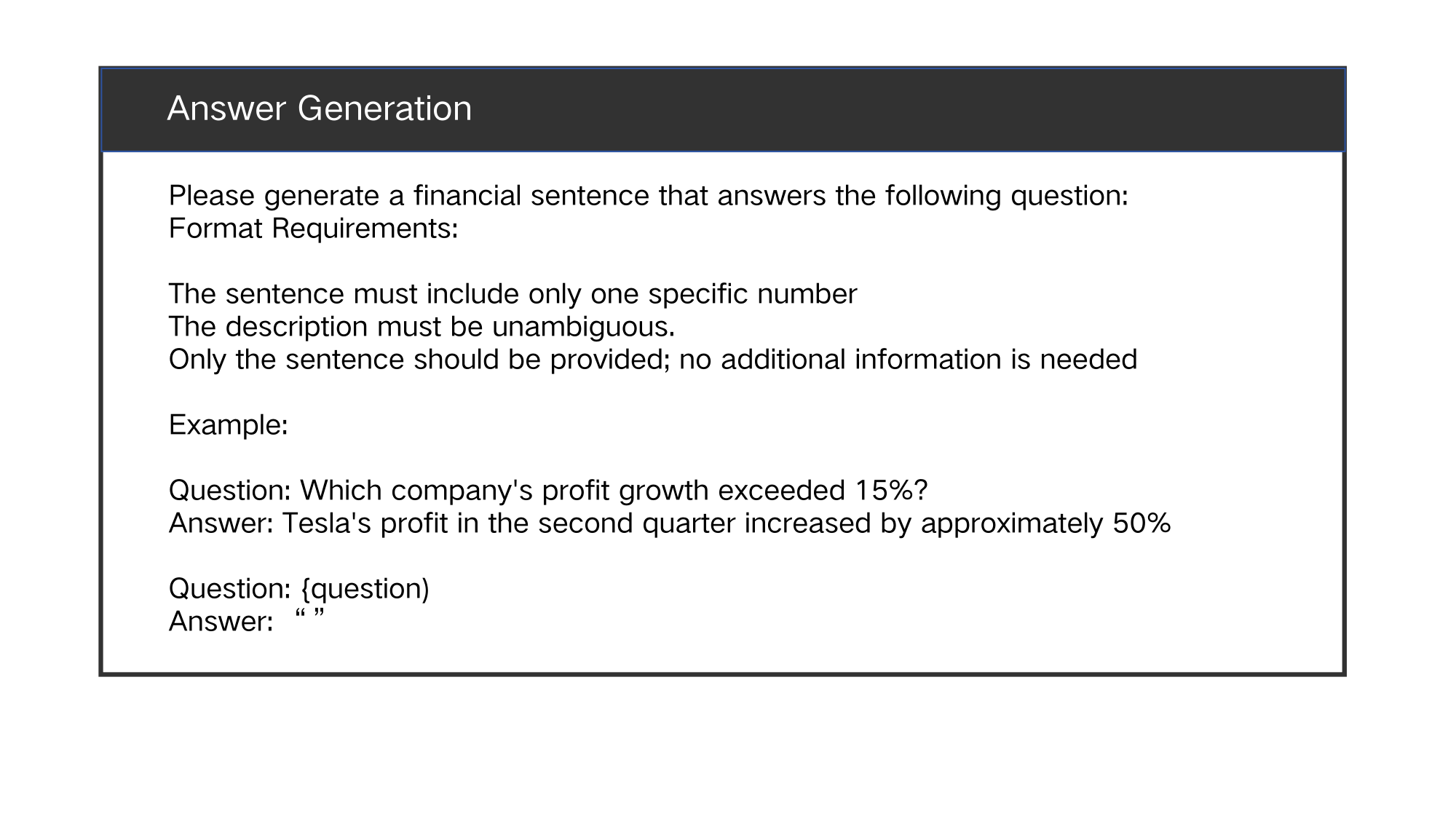}
\par\vspace{0pt}
Figure 10: Prompt used for answer generation.
\label{fig:answergenerate}
\end{minipage}

\end{document}